\documentclass[10pt,twocolumn,letterpaper]{article}
\usepackage[pagenumbers]{cvpr}
\usepackage{graphicx}
\usepackage{amsmath,amssymb,amsthm}
\usepackage{booktabs}
\usepackage{multirow}
\usepackage{xcolor}
\usepackage{bm}
\newtheorem{proposition}{Proposition}

\newtheorem{assumption}{Assumption}

\newtheorem{definition}{Definition}
\newtheorem*{proposition*}{Proposition}
\theoremstyle{remark}

\newcommand{\R}{\mathbb{R}}
\newcommand{\E}{\mathbb{E}}
\newcommand{\gq}{\nabla_{a} Q}
\newcommand{\Haa}{\nabla^2_{aa} Q}
\newcommand{\Has}{\nabla^2_{as} Q}
\newcommand{\npr}{POISE}
\usepackage{caption}
\usepackage{rotating}
\usepackage{placeins}
\makeatletter\renewcommand\LenToUnit[1]{\strip@pt\dimexpr(#1)*\p@/\unitlength\relax}\makeatother
\usepackage{array}
\usepackage{pifont}

\usepackage{colortbl}
\definecolor{good}{HTML}{1E9E5A}
\definecolor{bad}{HTML}{D84A4A}

\usepackage{algorithm}
\usepackage[noend]{algpseudocode}
\algrenewcommand\algorithmiccomment[1]{\hfill{\color{gray}$\triangleright$ #1}}

\newcommand{\gs}{\Delta\hat a}
\newcommand{\ga}{\bar a}

\definecolor{cvprblue}{rgb}{0.21,0.49,0.74}
\usepackage[pagebackref,breaklinks,colorlinks,allcolors=cvprblue]{hyperref}
\usepackage[capitalize]{cleveref}
\crefname{proposition}{Prop.}{Props.}
\crefname{assumption}{Assumption}{Assumptions}
\crefname{lemma}{Lemma}{Lemmas}
\crefname{corollary}{Cor.}{Cors.}
\crefname{algorithm}{Alg.}{Algs.}
\AtBeginDocument{\crefname{table}{Table}{Tables}}
\AtBeginDocument{\crefname{figure}{Figure}{Figures}}

\title{POISE: Curvature-Aware Critic Regularization\\for Smooth Actor--Critic Control}
\author{Jeong Woon Lee$^{1}$ \qquad Namho Kim$^{1}$ \qquad Hyoseok Hwang$^{1\dagger}$\\[2pt]
{\normalsize $^{1}$College of Software, Kyung Hee University}\\
{\tt\small \{everyman123, shawnbest, hyoseok\}@khu.ac.kr}\\
{\small $^{\dagger}$Corresponding author}}

\begin{document}
\maketitle
\begin{abstract}
Actor-critic methods achieve high returns in robotic control, yet trained actors exhibit high-frequency action oscillations.
Most smoothness methods constrain the actor, whereas PAVE regularizes the critic by separately penalizing action-gradient variation and encouraging negative action curvature.
The same gradient change, however, moves the critic's preferred action farther where the critic is flat, and separate penalties do not measure this motion.
We characterize this motion through the \emph{greedy shift}, which approximates greedy-action displacement across state transitions to first order under local concavity.
Since displacement alone does not reflect its critic-value consequences, the analysis further shows that the shift's curvature-weighted energy approximates the value lost by retaining the previous greedy action to second order.
Based on this relationship, \npr{} regularizes the curvature-weighted energy along temporal and spatial state displacements.
It employs a saddle-free curvature metric and reward-scale-consistent scaling while leaving the actor unchanged.
On six continuous control tasks with TD3, \npr{} reduces the smoothness score on every task by $5.9\times$ on average ($4.5\times$ geometric mean), compared with $2.2\times$ ($1.9\times$) for the strongest baseline, and achieves the lowest score on five tasks.
Paired return intervals include zero on four tasks, with a 14\% increase on Ant and an 11\% decrease on Hopper.
\end{abstract}

\section{Introduction}
\label{sec:intro}
\begin{figure}[t]
\centering
\includegraphics[width=\linewidth]{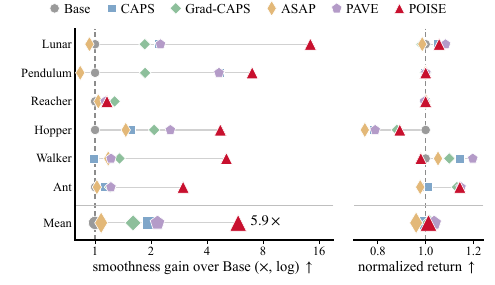}
\caption{\npr{} gives the largest smoothness gain over Base on five of six continuous control tasks with TD3. Each row is one task, and the last row is the six-task mean, $5.9\times$ for \npr{} against $2.2\times$ for PAVE, the strongest baseline. The right panel shows the return, normalized per task between a uniform-random policy at 0 and Base at 1.}
\label{fig:bench}
\end{figure}
Deep reinforcement learning (RL) has produced locomotion~\cite{hwangbo2019learning} and manipulation~\cite{levine2016end} controllers on real robots, and the off-policy actor-critic methods TD3~\cite{fujimoto2018addressing} and SAC~\cite{haarnoja2018soft} have become a standard choice for continuous control.
High task returns, however, do not guarantee smooth control: trained actors exhibit high-frequency action oscillations~\cite{mysore2021regularizing,christmann2024benchmarking}, which accelerate mechanical wear~\cite{inman_vibration,DELAPRESILLA2023108805} and hinder deployment in physical systems~\cite{mysore2021regularizing,song2023lipsnet}.

Most existing smoothness methods focus on the actor.
Some penalize action differences, or their changes, across consecutive or neighboring states~\cite{mysore2021regularizing,kobayashi2022l2c2,lee2024gradient}, others align actions with predictions derived from preceding transitions~\cite{kwak2026enhancingcontrolpolicysmoothness}, and architectural approaches constrain the sensitivity of the policy network to its inputs~\cite{song2023lipsnet,song2025lipsnet,takase2022stability}.
These methods limit variations in the actor's output and do not directly regularize the critic's action-gradient field, which continues to guide the actor during policy improvement.

A more direct approach is to regularize the critic, which provides the learning signal that guides the actor.
While earlier critic regularization methods primarily targeted learning stability and robustness, PAVE~\cite{lee2026stabilizing} established a connection between the critic's differential geometry and the sensitivity of its induced greedy policy.
PAVE characterizes this sensitivity through two geometric factors, the change in the action gradient across states and the local action curvature of the Q-function, and regularizes them separately.
However, the two factors jointly determine greedy-action motion.
The same gradient change induces a large shift along a flat direction but a small shift along a sharply curved direction, while a Euclidean penalty on the gradient change assigns equal costs to both.
Regularizing the two factors separately therefore does not directly regularize their combined effect on greedy-action motion.

We address this limitation by analyzing how the critic's preferred action moves across state transitions.
From the first-order optimality condition of the previous greedy action, we derive the \emph{greedy shift}, which combines the change in the action gradient with the inverse action curvature.
Under local concavity and a sufficiently small transition along the same branch of maximizers, the greedy shift approximates greedy-action displacement to first order.
However, action displacement alone does not account for its consequences in the critic's value landscape.
We therefore analyze the critic-value loss incurred by retaining the previous greedy action after a state transition, termed \emph{hold-regret}.
We establish that this loss is approximated to second order by half the curvature-weighted energy of the greedy shift.
Together, these results identify a single energy that captures the combined effect of gradient variation and directional curvature while assigning a local critic-value cost to greedy-action motion.

In this work, we propose \npr{} (POlicy-Induced Shift in the critic's curvature mEtric), a critic-side regularization method that penalizes the curvature-weighted energy derived from our analysis.
Since the theoretical energy assumes local concavity, \npr{} employs a saddle-free curvature metric to extend the formulation to replayed actions where this assumption does not necessarily hold.
With this metric, the regularizer evaluates action-gradient changes along observed state transitions and randomly reflected state displacements, extending regularization around the data without an additional perturbation-scale parameter.
The resulting energies are combined into a single penalty with curvature-dependent scaling, matching the reward-scaling behavior of the temporal-difference loss.
\npr{} integrates this single penalty with one weight into critic learning while leaving the actor unchanged.

On six continuous control tasks under a unified training protocol (\cref{fig:bench}), \npr{} with TD3 reduces the smoothness score on all tasks by $5.9\times$ on average ($4.5\times$ geometric mean), compared with $2.2\times$ ($1.9\times$) for PAVE, and achieves the lowest score on five tasks.
The paired return intervals include zero on four tasks, with a 14\% increase on Ant and an 11\% decrease on Hopper.

The main contributions are summarized as follows:
\begin{itemize}
\item We analyze greedy-action motion across state transitions and derive the greedy shift as its first-order approximation. We further establish a second-order relationship between its curvature-weighted energy and the local critic-value loss of retaining the previous greedy action.
\item We formulate \npr{}, a single critic-side regularizer that jointly accounts for gradient variation and directional curvature through a saddle-free metric and a reward-scale-consistent formulation, without modifying the actor.
\item We evaluate \npr{} on six continuous control tasks with TD3 and SAC, demonstrating a $5.9\times$ average reduction in the TD3 smoothness score, the largest among the actor-side and critic-side baselines. Critic diagnostics further examine the theoretical relationships under their local applicability conditions.
\end{itemize}

\section{Related Work}
\label{sec:related}
\subsection{Actor-side smoothness regularization}
\label{sec:rel_actor}
Penalty-based methods augment the actor's objective with terms that limit how much its output changes.
CAPS~\cite{mysore2021regularizing} penalizes action differences between consecutive states and between a state and Gaussian-perturbed neighbors.
L2C2~\cite{kobayashi2022l2c2} narrows the spatial neighborhood using the displacement between consecutive states and constrains the value function in the same way, on its output rather than its action gradient. Grad-CAPS~\cite{lee2024gradient} penalizes changes of consecutive action differences, and ASAP~\cite{kwak2026enhancingcontrolpolicysmoothness} aligns each action with the action predicted from the preceding state.
Other methods bound the Lipschitz constant of the policy network, through spectral normalization of a stability-certified controller~\cite{takase2022stability}, a gradient penalty on the policy~\cite{chen2024lcp}, gradient normalization in LipsNet~\cite{song2023lipsnet}, or Jacobian regularization with a Fourier filter layer in LipsNet++~\cite{song2025lipsnet}, and action chunking with temporal ensembling in imitation learning~\cite{zhao2023learning} smooths the executed action sequence.
CATS~\cite{ahn2026revisiting} shows that a temporal penalty on observed transitions also bounds the expected action difference between states that share a next state, and ramps the penalty up linearly during training.
Apart from the value constraint of L2C2, all of these constrain the outputs or the architecture of the policy; \npr{} instead regularizes the action-gradient field of the critic.

\subsection{Critic geometry and gradient regularization}
\label{sec:rel_critic}
Previous critic regularization methods primarily address the stability and robustness of value learning rather than the motion of the induced policy.
Gradient and Jacobian penalties have been used to improve the generalization~\cite{sokolic2017robust} and the training stability~\cite{roth2017stabilizing} of learned functions, while spectral normalization controls network sensitivity, applied to the value network~\cite{gogianu2021spectral} or to both actor and critic~\cite{bjorck2021towards}.
SR$^2$L~\cite{shen2020deep} also introduces critic-side regularization for policy smoothness, but constrains changes in $Q$-values under bounded state perturbations rather than the critic's action-gradient geometry.
PAVE~\cite{lee2026stabilizing} explicitly connects the critic's differential geometry to greedy-policy sensitivity and regularizes the action-gradient field and action curvature with separate penalties, Euclidean terms on the gradient field and a hinge with a fixed margin on the curvature.
However, these penalties do not directly account for how the two factors jointly determine greedy-action motion.
\npr{} instead regularizes their combined effect, the greedy shift, in the critic's curvature metric.

\subsection{Second-order structure in continuous control}
\label{sec:rel_second}
Second-order information usually enters RL as an optimizer: natural-gradient methods precondition the actor's parameter gradient with the Fisher information~\cite{kakade2001natural}, and saddle-free Newton methods replace the Hessian by its absolute value to escape saddle points~\cite{dauphin2014identifying}.
Value-gradient learning~\cite{fairbank2012value} trains the gradient of the value function directly, since a greedy policy depends on the value only through its gradient, and quadratic critics~\cite{gu2016continuous} make the greedy action available in closed form.
\npr{} uses the action Hessian in neither role: it is neither an optimizer nor a parameterization, but the metric in which the motion of the greedy action is measured and penalized, and its parameters are trained with the same first-order optimizer as every baseline.

\section{Preliminaries}
\label{sec:prelim}
\paragraph{Actor-critic control.}
We consider an MDP $(\mathcal S,\mathcal A,P,r,\gamma)$ with states $s\in\mathcal S\subset\R^k$ and actions $a\in\mathcal A\subset\R^d$.
Off-policy actor-critic methods (DDPG~\cite{lillicrap2015continuous}, TD3~\cite{fujimoto2018addressing}) learn a critic $Q_\theta$ by temporal-difference (TD) regression on replayed transitions $(s,a,r,s')\sim\mathcal D$ and a deterministic actor $\pi_\phi$ by ascending the critic~\cite{silver2014deterministic},
\begin{align}
\mathcal L_{\mathrm{TD}}(\theta)&=\E_{(s,a,r,s')\sim\mathcal D}\big[(Q_\theta(s,a)-y)^2\big],\label{eq:td}\\
\nabla_\phi J(\phi)&=\E_{s}\big[\gq_\theta(s,a)\big|_{a=\pi_\phi(s)}\,\nabla_\phi\pi_\phi(s)\big],\label{eq:dpg}
\end{align}
where $y$ is the algorithm-specific TD target (twin-critic minimum and target smoothing for TD3, entropy term for SAC) and \cref{eq:td} is written per critic head.
The actor update is guided by the critic's action gradient, which motivates the study of the greedy policy induced by the critic.
We consider the greedy policy $a^*(s)$, a locally unique interior maximizer of $Q(s,\cdot)$ that varies smoothly with the state, characterized by
\begin{equation}
\gq\big(s,a^*(s)\big)=0,\qquad \Haa\big(s,a^*(s)\big)\prec0 .
\label{eq:greedy}
\end{equation}
Throughout,
\begin{equation}
\begin{aligned}
g&=\gq\in\R^d,\qquad C=\Has\in\R^{d\times k},\\
B&=-\Haa\in\R^{d\times d},
\end{aligned}
\label{eq:notation}
\end{equation}
denote the action gradient, the mixed state-action derivatives and the negative action Hessian; the analysis below assumes $B\succ0$.

\paragraph{Action smoothness.}
The smoothness of a deployed policy is measured, as in~\cite{mysore2021regularizing,christmann2024benchmarking}, on its action sequence $(a_t)_{t=0}^{T-1}$ by the smoothness score
\begin{equation}
\mathrm{sm}=\frac{2}{T^2}\sum_{f=1}^{\lfloor T/2\rfloor-1} f\,|\hat a_f|,\qquad \hat a_f=\sum_{t=0}^{T-1}a_t\,e^{-2\pi i ft/T},
\label{eq:sm}
\end{equation}
with $f$ in cycles per episode: the frequency-weighted magnitude of the action spectrum. The score is computed for each action dimension and averaged across dimensions; lower values indicate smoother action sequences.
It is a temporal quantity along trajectories; the regularizers that target it act either on the actor's output or on the critic.

\paragraph{The critic-geometry view.}
At a greedy action satisfying \cref{eq:greedy}, differentiating the optimality condition with respect to the state gives
\begin{equation}
\nabla_s a^*(s)=B^{-1}C,\qquad \|\nabla_s a^*(s)\|_2\le\frac{\|C\|_2}{\lambda_{\min}(B)},
\label{eq:ift}
\end{equation}
so the spatial sensitivity of the greedy policy is bounded by two factors of the critic's geometry: the mixed derivative $C$ in the numerator and the smallest action curvature $\lambda_{\min}(B)$ in the denominator~\cite{lee2026stabilizing}.
PAVE~\cite{lee2026stabilizing} regularizes each factor on the critic with its own weight: two Euclidean penalties on the change of the action gradient $g$ across perturbed and consecutive states lower the numerator, and a hinge on the action curvature along sampled directions raises the denominator.
PAVE established a connection between the critic's differential geometry and the sensitivity of its induced greedy policy.
The bound limits the worst case over all state directions.
Along an observed transition, the same change in the action gradient moves the greedy action farther where the action curvature is small, and a penalty on the gradient change alone assigns both cases the same cost.
Our work therefore analyzes the \emph{temporal} motion of the greedy action across a transition directly.

\section{How Critic Geometry Moves the Greedy Action}
\label{sec:theory}
The local sensitivity bound does not directly quantify the displacement of the critic's preferred action along a particular state transition.
This motivates a transition-level analysis of greedy-action motion and its associated local critic-value cost.

\subsection{Assumptions and the greedy shift}
\label{sec:theory_def}
Let $\Delta s=s'-s$, $\ga=a^*(s)$ and $\Delta a^*=a^*(s')-\ga$.
The change of the action gradient across the transition is $u=g(s',\ga)-g(s,\ga)$.
\begin{assumption}[Local concavity]\label{ass:A}
$Q$ is three times continuously differentiable, and $\ga$ is an interior local maximizer of $Q(s,\cdot)$ with $B(s,\ga)\succ0$.
\end{assumption}
\begin{assumption}[Local transition]\label{ass:B}
The next state $s'$ lies in a neighborhood of $s$ in which $a^*(s')$ stays on the smooth branch of maximizers through $\ga$.
\end{assumption}
Under \cref{ass:A,ass:B} the implicit function theorem and the smoothness of $Q$ give $\Delta a^*=O(\|\Delta s\|)$ and $B(s',\ga)=B(s,\ga)+O(\|\Delta s\|)$ as $\Delta s\to0$.
Every $O(\cdot)$ term in this section refers to this limit.

The greedy action at $s'$ is the point where the action gradient vanishes again, $g(s',\ga+\Delta a^*)=0$.
Since $\ga$ maximizes $Q(s,\cdot)$ and $\nabla_a g=-B$, a first-order expansion of $g(s',\cdot)$ in the action around $\ga$ gives
\begin{align*}
0&=g(s',\ga+\Delta a^*)\\
&=g(s',\ga)+\nabla_a g(s',\ga)\,\Delta a^*+O(\|\Delta a^*\|^2)\\
&=\big[g(s',\ga)-\underbrace{g(s,\ga)}_{=0}\big]-B(s',\ga)\,\Delta a^*+O(\|\Delta a^*\|^2)\\
&=u-B(s',\ga)\,\Delta a^*+O(\|\Delta a^*\|^2).
\end{align*}
The two relations above turn the last line into $B(s,\ga)\,\Delta a^*=u+O(\|\Delta s\|^2)$.
Since $B(s,\ga)$ is invertible under \cref{ass:A},
\begin{equation}
\Delta a^*=B(s,\ga)^{-1}u+O(\|\Delta s\|^2).
\label{eq:increment}
\end{equation}
The leading term is computed from derivatives of the critic alone, without solving the maximization at $s'$.
\begin{definition}[Greedy shift]\label{def:gs}
For a transition from $s$ to $s'$ and an action $a$ at which $B(s,a)$ is invertible, the greedy shift is
\begin{equation}
\gs=B(s,a)^{-1}u,\qquad u=g(s',a)-g(s,a).
\label{eq:ux}
\end{equation}
\end{definition}
By \cref{eq:increment}, the greedy shift at $a=\ga$ measures how far the critic's preferred action moves across the transition, up to a remainder of second order in $\|\Delta s\|$; \cref{sec:method} extends the construction beyond this regime as a training surrogate.

\subsection{Value cost of the greedy shift}
\label{sec:theory_cost}
\begin{proposition}[Hold-regret]\label{cor:regret}
Under \cref{ass:A,ass:B} with $u$ and $B$ evaluated at $a=\ga$, the critic value lost by retaining $\ga$ at $s'$ is
\begin{equation}
Q(s',a^*(s'))-Q(s',\ga)=\tfrac12 u^\top B^{-1}u+O(\|\Delta s\|^3).
\label{eq:regret}
\end{equation}
\end{proposition}
\begin{proof}
The lost value compares two actions on the same critic $Q(s',\cdot)$: the new maximizer $a^*(s')$ and the retained action $\ga=a^*(s')-\Delta a^*$.
Since $a^*(s')$ maximizes $Q(s',\cdot)$ and $\nabla^2_{aa}Q=-B$, a second-order expansion of $Q(s',\cdot)$ in the action around $a^*(s')$ gives
\begin{align*}
Q(s',\ga)&=Q\big(s',a^*(s')-\Delta a^*\big)\\
&=Q\big(s',a^*(s')\big)-\underbrace{g\big(s',a^*(s')\big)^\top}_{=0}\Delta a^*\\
&\quad-\tfrac12\Delta a^{*\top}B\big(s',a^*(s')\big)\Delta a^*+O(\|\Delta a^*\|^3).
\end{align*}
\Cref{ass:A,ass:B} give $B(s',a^*(s'))=B+O(\|\Delta s\|)$ and $u=O(\|\Delta s\|)$ with $B=B(s,\ga)$.
\Cref{eq:increment} gives $\Delta a^*=\gs+O(\|\Delta s\|^2)$ with $\gs=B^{-1}u$, so
\begin{align*}
&Q\big(s',a^*(s')\big)-Q(s',\ga)\\
&\quad=\tfrac12\Delta a^{*\top}B\,\Delta a^*+O(\|\Delta s\|^3)\\
&\quad=\tfrac12\gs^\top B\,\gs+O(\|\Delta s\|^3)\\
&\quad=\tfrac12u^\top B^{-1}u+O(\|\Delta s\|^3).\qedhere
\end{align*}
\end{proof}

The greedy shift is one Newton step on $Q(s',\cdot)$ from $\ga$ to leading order, and the hold-regret is half the squared Newton decrement to the same order~\cite{boyd2004convex}.
Both quantities describe the critic's greedy action, not the executed actor or the environment return.

\subsection{Anisotropy of the greedy shift}
\label{sec:theory_aniso}
The greedy shift and the hold-regret depend on $u$ and $B$ together, whereas the Euclidean penalties of PAVE measure $\|u\|^2$ alone.
With several action dimensions, $B$ has a different curvature along each direction, and the greedy shift divides the component of $u$ along each direction by the curvature of that direction.
\begin{proposition}[Anisotropy]\label{prop:gap}
Let $\lambda_{\min}$ and $\lambda_{\max}$ be the smallest and largest eigenvalues of $B\succ0$, the curvatures along the flattest and stiffest action directions. Then
\begin{equation}
\begin{aligned}
\frac{\|u\|}{\lambda_{\max}}&\le\|\gs\|\le\frac{\|u\|}{\lambda_{\min}},\\
\frac{\|u\|^2}{\lambda_{\max}}&\le u^\top B^{-1}u\le\frac{\|u\|^2}{\lambda_{\min}},
\end{aligned}
\label{eq:gap}
\end{equation}
with equality when $u$ lies along the stiffest and the flattest eigenvector, respectively.
\end{proposition}
\begin{proof}
In the eigenbasis of $B$, $\|\gs\|^2=\sum_iu_i^2/\lambda_i^2$ and $u^\top B^{-1}u=\sum_iu_i^2/\lambda_i$ with $\lambda_{\min}\le\lambda_i\le\lambda_{\max}$.
\end{proof}
Two gradient changes of the same norm therefore produce greedy shifts and value costs that differ by a factor of up to $\mathrm{cond}(B)=\lambda_{\max}/\lambda_{\min}$, and a Euclidean penalty $\|u\|^2$ assigns both the same cost.

\subsection{One quantity: the curvature-weighted energy}
\label{sec:theory_energy}
The results above meet in one scalar, the curvature-weighted energy
\begin{equation}
\begin{aligned}
u^\top B^{-1}u&=\gs^\top B\,\gs\\
&=2\big[Q(s',a^*(s'))-Q(s',\ga)\big]+O(\|\Delta s\|^3),
\end{aligned}
\label{eq:energy}
\end{equation}
the motion of the preferred action measured in units of critic value.
For a fixed positive-definite $B$, the energy controls the squared greedy shift at least as tightly as the gradient change alone does,
\begin{equation}
\|\gs\|^2\le\frac{u^\top B^{-1}u}{\lambda_{\min}}\le\frac{\|u\|^2}{\lambda_{\min}^2}.
\label{eq:energy_bound}
\end{equation}
The smallest eigenvalue of $B$ changes during training, so a decrease of the energy does not guarantee less motion of a learned actor.
A penalty on $\|u\|^2$ omits the curvature, and a penalty on $\|\gs\|^2$ omits the critic-value weighting.
This distinction motivates the practical regularizer developed next.

\section{POISE: Curvature-Weighted Greedy Shift Regularization}
\label{sec:method}

\subsection{Saddle-free curvature metric}
\label{sec:method_metric}
The energy of \cref{sec:theory} assumes $B\succ0$, but at replayed actions $B$ has negative or near-zero eigenvalues wherever the critic is not concave.
Along a negative eigenvalue the energy is negative, so its minimization rewards non-concave curvature, and along a near-zero eigenvalue it is unbounded.
The metric therefore follows saddle-free Newton methods~\cite{dauphin2014identifying} and replaces each eigenvalue $\lambda_i$ of $B=V\mathrm{diag}(\lambda_i)V^\top$ by its absolute value with a floor,
\begin{equation}
\begin{aligned}
|B|&=V\,\mathrm{diag}\!\big(\max(|\lambda_i|,\eta\kappa)\big)V^\top\\
&=B+\mathrm{sg}\!\left(V\,\mathrm{diag}\!\big(\max(|\lambda_i|,\eta\kappa)-\lambda_i\big)V^\top\right),
\end{aligned}
\label{eq:absB}
\end{equation}
where $\kappa$ is the curvature scale of \cref{eq:npr}, $\eta>0$ sets the floor relative to it and $\mathrm{sg}$ stops gradients.
The second line has the same value and defines the semi-gradient, which flows through $B$ and not through the correction or $\kappa$.
For $\kappa>0$ the metric satisfies $|B|\succeq\eta\kappa I$, so $0\le u^\top|B|^{-1}u\le\|u\|^2/(\eta\kappa)$, and it coincides with $B$ where $B\succeq\eta\kappa I$.

\subsection{Temporal and spatial displacements}
\label{sec:method_disp}
The energy is evaluated on two state displacements of a replayed transition $(s,a,s')\sim\mathcal D$ with $\Delta s=s'-s$.
The temporal displacement $\Delta s$ probes the critic along the one direction that the data took.
The spatial displacement $\varepsilon\odot\Delta s$ with $\varepsilon\in\{\pm1\}^k$ samples independently sign-reflected versions of the observed state difference around $s$; the reflected states need not lie on the data manifold.
Unlike Gaussian noise~\cite{mysore2021regularizing,lee2026stabilizing}, the reflection needs no noise scale, since each coordinate keeps its size in the data.
The two gradient changes are
\begin{equation}
\begin{aligned}
u_{\mathrm T}&=g(s',a)-g(s,a),\\
u_{\mathrm S}&=g(s+\varepsilon\odot\Delta s,a)-g(s,a),
\end{aligned}
\label{eq:u}
\end{equation}
and the saddle-free inverse $|B|^{-1}$ in place of $B^{-1}$ in \cref{def:gs} gives the corresponding surrogate shifts at replayed actions,
\begin{equation}
\gs_{\mathrm T}=|B|^{-1}u_{\mathrm T},\qquad \gs_{\mathrm S}=|B|^{-1}u_{\mathrm S}.
\label{eq:x}
\end{equation}

\subsection{Curvature-weighted loss}
\label{sec:method_loss}
The \npr{} loss sums the energies of the two shifts and scales them by $\kappa$,
\begin{equation}
\begin{aligned}
\mathcal L_{\mathrm{POISE}}&=\kappa\,\E_{\mathcal D}\!\left[u_{\mathrm T}^\top|B|^{-1}u_{\mathrm T}+u_{\mathrm S}^\top|B|^{-1}u_{\mathrm S}\right],\\
\kappa&=\mathrm{sg}\Big(\E_{\mathcal D}\!\Big[\tfrac1d\textstyle\sum_{i=1}^{d}|\lambda_i(B)|\Big]\Big).
\end{aligned}
\label{eq:npr}
\end{equation}
The scale $\kappa$ is the detached mean absolute curvature of the critic and is not a hyperparameter.
When $Q$ and the rewards are scaled by $c>0$, the energy and $\kappa$ each scale by $c$, so for $\kappa>0$ the loss scales as $c^2$, as the TD loss of TD3 does and as that of SAC does when its entropy temperature scales by $c$.
At the greedy action $\ga$ with $B\succeq\eta\kappa I$, the temporal term equals $2\kappa$ times the hold-regret of \cref{cor:regret} to leading order.

\subsection{Total objective}
\label{sec:method_signal}
The objective for the critic parameters $\theta$ is the weighted sum of the temporal-difference error and the \npr{} loss,
\begin{equation}
\mathcal L(\theta)=\mathcal L_{\mathrm{TD}}+\lambda\,\mathcal L_{\mathrm{POISE}},\qquad \lambda=\tilde\lambda/d,
\label{eq:total}
\end{equation}
where $\tilde\lambda$ is the single weight of the method, $\mathcal L_{\mathrm{TD}}$ is the standard temporal-difference objective and $d$ is the action dimension.
The loss acts on the critic alone, and the actor update is unchanged.

\begin{table*}[t]
\centering
\setlength{\tabcolsep}{3.4pt}
\renewcommand{\arraystretch}{0.95}
\caption{Quantitative comparison of cumulative return (re) and smoothness score (sm) under TD3 and SAC (5 seeds, standard deviation in parentheses). Higher re and lower sm are better. \textbf{Bold}: best; \underline{underline}: second. \textcolor{red}{Red}: paired 95\% bootstrap interval of the return difference to Base entirely below zero.}
\label{tab:main}
\scriptsize
\begin{tabular}{@{}lcccccccccccc@{}}
\toprule
& \multicolumn{2}{c}{Lunar} & \multicolumn{2}{c}{Pendulum} & \multicolumn{2}{c}{Reacher} & \multicolumn{2}{c}{Hopper} & \multicolumn{2}{c}{Walker} & \multicolumn{2}{c}{Ant}\\
\cmidrule(lr){2-3} \cmidrule(lr){4-5} \cmidrule(lr){6-7} \cmidrule(lr){8-9} \cmidrule(lr){10-11} \cmidrule(lr){12-13}
Method & re$\uparrow$ & sm$\downarrow$ & re$\uparrow$ & sm$\downarrow$ & re$\uparrow$ & sm$\downarrow$ & re$\uparrow$ & sm$\downarrow$ & re$\uparrow$ & sm$\downarrow$ & re$\uparrow$ & sm$\downarrow$\\
\midrule
TD3 Base & 214.0{\tiny\,(65.3)} & 1.580{\tiny\,(0.571)} & -154.5{\tiny\,(0.78)} & 1.900{\tiny\,(0.381)} & \underline{-3.36}{\tiny\,(0.08)} & 0.051{\tiny\,(0.001)} & \textbf{3595}{\tiny\,(58.3)} & 2.645{\tiny\,(0.363)} & 4596{\tiny\,(562.3)} & 1.818{\tiny\,(0.307)} & 4856{\tiny\,(1008)} & 2.158{\tiny\,(0.174)}\\
\multicolumn{13}{l}{\cellcolor{gray!12}\textit{Actor-side regularization}}\\
CAPS~\cite{mysore2021regularizing} & 236.2{\tiny\,(64.0)} & 0.715{\tiny\,(0.292)} & \textcolor{red}{-160.5}{\tiny\,(5.35)} & \underline{0.407}{\tiny\,(0.029)} & -3.37{\tiny\,(0.07)} & 0.045{\tiny\,(0.002)} & \textcolor{red}{2822}{\tiny\,(1566)} & 1.706{\tiny\,(0.628)} & \underline{5259}{\tiny\,(679.7)} & 1.848{\tiny\,(0.280)} & 4912{\tiny\,(462.4)} & 1.922{\tiny\,(0.068)}\\
Grad-CAPS~\cite{lee2024gradient} & 205.9{\tiny\,(53.0)} & 0.855{\tiny\,(0.147)} & -155.1{\tiny\,(2.28)} & 1.023{\tiny\,(0.519)} & \underline{-3.36}{\tiny\,(0.04)} & \textbf{0.040}{\tiny\,(0.001)} & 3164{\tiny\,(677.0)} & 1.273{\tiny\,(0.387)} & 5049{\tiny\,(694.8)} & \underline{1.344}{\tiny\,(0.314)} & 5500{\tiny\,(403.5)} & 1.793{\tiny\,(0.152)}\\
ASAP~\cite{kwak2026enhancingcontrolpolicysmoothness} & 207.5{\tiny\,(46.3)} & 1.694{\tiny\,(0.676)} & -154.9{\tiny\,(2.95)} & 2.283{\tiny\,(0.906)} & \textbf{-3.32}{\tiny\,(0.02)} & 0.049{\tiny\,(0.003)} & \textcolor{red}{2685}{\tiny\,(849.5)} & 1.809{\tiny\,(0.394)} & 4834{\tiny\,(540.5)} & 1.544{\tiny\,(0.221)} & 4744{\tiny\,(683.2)} & 2.110{\tiny\,(0.187)}\\
\multicolumn{13}{l}{\cellcolor{gray!12}\textit{Critic-side regularization}}\\
PAVE~\cite{lee2026stabilizing} & \textbf{252.0}{\tiny\,(19.6)} & \underline{0.703}{\tiny\,(0.269)} & \textbf{-153.6}{\tiny\,(0.86)} & 0.410{\tiny\,(0.025)} & \textcolor{red}{-3.57}{\tiny\,(0.14)} & 0.045{\tiny\,(0.004)} & 2840{\tiny\,(968.8)} & \underline{1.042}{\tiny\,(0.306)} & \textbf{5501}{\tiny\,(512.6)} & 1.492{\tiny\,(0.378)} & \textbf{5583}{\tiny\,(413.9)} & \underline{1.774}{\tiny\,(0.120)}\\
\rowcolor{blue!8}\textbf{\npr{} (ours)} & \underline{239.8}{\tiny\,(33.9)} & \textbf{0.110}{\tiny\,(0.011)} & \underline{-154.4}{\tiny\,(0.55)} & \textbf{0.272}{\tiny\,(0.017)} & \underline{-3.36}{\tiny\,(0.04)} & \underline{0.044}{\tiny\,(0.001)} & \textcolor{red}{\underline{3208}}{\tiny\,(623.1)} & \textbf{0.561}{\tiny\,(0.101)} & 4503{\tiny\,(621.8)} & \textbf{0.358}{\tiny\,(0.082)} & \underline{5559}{\tiny\,(292.4)} & \textbf{0.726}{\tiny\,(0.092)}\\
\midrule
SAC Base & \underline{264.6}{\tiny\,(20.8)} & 0.318{\tiny\,(0.066)} & \underline{-149.3}{\tiny\,(1.18)} & 0.510{\tiny\,(0.052)} & \textbf{-3.46}{\tiny\,(0.03)} & 0.051{\tiny\,(0.002)} & \underline{3168}{\tiny\,(435.6)} & 0.778{\tiny\,(0.053)} & \textbf{4843}{\tiny\,(513.6)} & 0.712{\tiny\,(0.100)} & \underline{5380}{\tiny\,(435.7)} & 1.994{\tiny\,(0.109)}\\
\multicolumn{13}{l}{\cellcolor{gray!12}\textit{Actor-side regularization}}\\
CAPS~\cite{mysore2021regularizing} & 216.1{\tiny\,(122.6)} & 0.284{\tiny\,(0.028)} & \textcolor{red}{-151.8}{\tiny\,(1.55)} & \underline{0.316}{\tiny\,(0.016)} & -3.51{\tiny\,(0.11)} & \underline{0.045}{\tiny\,(0.003)} & 3041{\tiny\,(966.5)} & 0.599{\tiny\,(0.135)} & 4541{\tiny\,(580.1)} & 0.667{\tiny\,(0.052)} & 5288{\tiny\,(889.8)} & 1.906{\tiny\,(0.191)}\\
Grad-CAPS~\cite{lee2024gradient} & 215.9{\tiny\,(126.6)} & 0.236{\tiny\,(0.025)} & \textbf{-148.8}{\tiny\,(0.86)} & 0.402{\tiny\,(0.017)} & \textcolor{red}{-3.53}{\tiny\,(0.08)} & \textbf{0.043}{\tiny\,(0.003)} & 2914{\tiny\,(852.0)} & 0.574{\tiny\,(0.116)} & \underline{4754}{\tiny\,(386.1)} & 0.558{\tiny\,(0.035)} & 5237{\tiny\,(922.1)} & 1.803{\tiny\,(0.101)}\\
ASAP~\cite{kwak2026enhancingcontrolpolicysmoothness} & 221.1{\tiny\,(124.4)} & \textbf{0.188}{\tiny\,(0.015)} & -149.8{\tiny\,(0.93)} & 0.379{\tiny\,(0.055)} & \textcolor{red}{-3.51}{\tiny\,(0.03)} & \textbf{0.043}{\tiny\,(0.003)} & \textbf{3297}{\tiny\,(378.0)} & \underline{0.469}{\tiny\,(0.061)} & 4392{\tiny\,(734.4)} & \underline{0.529}{\tiny\,(0.059)} & 4659{\tiny\,(2422)} & \underline{1.520}{\tiny\,(0.205)}\\
\multicolumn{13}{l}{\cellcolor{gray!12}\textit{Critic-side regularization}}\\
PAVE~\cite{lee2026stabilizing} & 264.2{\tiny\,(12.5)} & \underline{0.200}{\tiny\,(0.015)} & \textcolor{red}{-151.0}{\tiny\,(0.64)} & 0.347{\tiny\,(0.025)} & \underline{-3.48}{\tiny\,(0.17)} & 0.051{\tiny\,(0.002)} & 3062{\tiny\,(863.2)} & 0.549{\tiny\,(0.090)} & \textcolor{red}{4545}{\tiny\,(388.9)} & 0.568{\tiny\,(0.068)} & \textbf{5502}{\tiny\,(384.1)} & 1.628{\tiny\,(0.126)}\\
\rowcolor{blue!8}\textbf{\npr{} (ours)} & \textbf{267.2}{\tiny\,(9.76)} & 0.221{\tiny\,(0.030)} & \textcolor{red}{-151.7}{\tiny\,(1.12)} & \textbf{0.286}{\tiny\,(0.019)} & \underline{-3.48}{\tiny\,(0.07)} & 0.047{\tiny\,(0.001)} & 2698{\tiny\,(647.4)} & \textbf{0.184}{\tiny\,(0.019)} & 4528{\tiny\,(522.1)} & \textbf{0.329}{\tiny\,(0.094)} & 5036{\tiny\,(112.5)} & \textbf{0.638}{\tiny\,(0.059)}\\
\bottomrule
\end{tabular}
\end{table*}

\section{Experiments}
\label{sec:exp}

\subsection{Experimental setup}
\label{sec:exp_setup}
We followed the experimental protocol of PAVE~\cite{lee2026stabilizing}.
The benchmark comprised the six Gymnasium~\cite{towers2024gymnasium} and MuJoCo~\cite{todorov2012mujoco} environments of PAVE: LunarLanderContinuous, Pendulum, Reacher, Hopper, Walker2d and Ant.
The metrics were the cumulative return re and the smoothness score sm of \cref{eq:sm}, for which lower is smoother.
The backbones were TD3~\cite{fujimoto2018addressing} and SAC~\cite{haarnoja2018soft} on Stable-Baselines3~\cite{raffin2021sb3}, and all methods used SiLU activations~\cite{elfwing2018sigmoid}.
The baselines were CAPS~\cite{mysore2021regularizing}, Grad-CAPS~\cite{lee2024gradient}, ASAP~\cite{kwak2026enhancingcontrolpolicysmoothness} and PAVE, each with the per-task hyperparameters published in~\cite{lee2026stabilizing}.
The weight of \npr{} in \cref{tab:main} was selected per task on the test runs, following the selection rule of PAVE.
A single shared weight fixed on disjoint validation seeds kept the lowest sm among the regularized methods on all six TD3 tasks and four SAC tasks, with a mean reduction against Base of $6.2\times$ on TD3 and $2.4\times$ on SAC and a return below that of the per-task weight on four TD3 tasks, as \cref{tab:shared} of the supplementary material reports.
The penalty was applied to the first critic in TD3, the critic that the actor ascends, and to both critics in SAC, where the two penalties were summed, with the floor $\eta=0.1$ on every task.
We retrained every method under this protocol with the same five seeds, evaluated ten test episodes per seed with shared reset seeds, and reported the mean and standard deviation over seeds.
The main results covered both backbones, and every further analysis used TD3.

\subsection{Main results}
\label{sec:exp_main}
\Cref{tab:main} evaluated whether \npr{} reduced the smoothness score while retaining the return of the unregularized Base, in comparison with actor-side and critic-side regularization.
For TD3, actor-side regularization yielded limited and inconsistent smoothing: the mean reduction of the Base score remained below $2\times$ for all three methods, and CAPS and ASAP increased the score on one and two tasks, respectively.
Critic-side regularization was more effective, and the proposed method improved substantially on PAVE, with a mean reduction of $5.9\times$ against $2.2\times$ and the lowest score on five of the six tasks.
The only exception was Reacher, where the Base policy was already smooth and Grad-CAPS attained 0.040 against 0.044 for our method.
Hopper was the only task marked in red for \npr{}: the mean return decreased by 11\% on Hopper and increased by 14\% on Ant.
Apart from Grad-CAPS on Reacher, every baseline that returned more than \npr{} was less smooth: on Walker, all four baselines returned more, with smoothness scores 3.8 to 5.2 times that of \npr{}.

In SAC, the Base policy was smoother than under TD3 on five tasks, and the reductions achieved by CAPS, Grad-CAPS, PAVE and \npr{} were smaller.
Both observations are consistent with the objectives of the two actors: the TD3 actor ascends a single critic and tracks the greedy action that \npr{} regularizes, whereas the SAC actor is trained toward a distribution proportional to $\exp(Q/\alpha)$ for the minimum $Q$ of two critics and the entropy temperature $\alpha$.
The mean of this target distribution varies continuously where the greedy action jumps between ridges.
The relative order of the two families changed: the actor-side ASAP attained a larger mean reduction than the critic-side PAVE, $1.4\times$ against $1.3\times$, whereas the proposed method retained the largest reduction, $2.3\times$, and the lowest score on Pendulum, Hopper, Walker and Ant.
On Lunar and Reacher, where the SAC Base was already smooth, ASAP attained the lowest scores, 0.188 and 0.043, the second tied with Grad-CAPS, against 0.221 and 0.047 for \npr{}.
The mean return of our method was 15\% below Base on Hopper, 6.5\% below on Walker and 6.4\% below on Ant, and the only decrease marked in red was on Pendulum, from $-149.3$ to $-151.7$.
Every baseline that returned more than \npr{} was again less smooth: on Hopper, all four baselines returned more, with scores 2.6 to 3.3 times that of \npr{}.

\begin{table}[t]
\centering
\setlength{\tabcolsep}{4.2pt}
\caption{Design analysis of \npr{} (TD3, 5 seeds, shared weight). Each row above \npr{} replaces one design choice by its simpler alternative. \textbf{Bold}: best; \underline{underline}: second.}
\label{tab:ablation}
\scriptsize
\begin{tabular}{@{}lcccccc@{}}
\toprule
& \multicolumn{2}{c}{Lunar} & \multicolumn{2}{c}{Hopper} & \multicolumn{2}{c}{Walker}\\
\cmidrule(lr){2-3}\cmidrule(lr){4-5}\cmidrule(lr){6-7}
Variant & re$\uparrow$ & sm$\downarrow$ & re$\uparrow$ & sm$\downarrow$ & re$\uparrow$ & sm$\downarrow$\\
\midrule
Base & 214.0 & 1.580 & \textbf{3595} & 2.645 & \underline{4596} & 1.818\\
\midrule
Curvature metric detached & \textbf{227.5} & 1.902 & 2673 & 3.050 & \textbf{4982} & 1.699\\
Temporal displacement only & 219.6 & 0.279 & 2305 & 0.333 & 4517 & 0.517\\
Squared shift $\kappa^2\|\gs\|^2$ & 211.6 & \textbf{0.167} & \underline{2783} & \textbf{0.122} & 2583 & \textbf{0.217}\\
\rowcolor{blue!8}\npr{} (ours) & \underline{226.0} & \underline{0.205} & 2672 & \underline{0.204} & 4503 & \underline{0.358}\\
\bottomrule
\end{tabular}
\end{table}

\subsection{Design analysis}
\label{sec:exp_abl}
\Cref{tab:ablation} replaced one design choice of \npr{} at a time by its simpler alternative at a single shared weight $\lambda=0.1/d$ for action dimension $d$, fixed on disjoint validation seeds.
The gradient through the curvature metric was necessary: with $|B|$ detached, the score stayed at the level of Base on all three tasks, 1.90, 3.05 and 1.70 against 1.58, 2.65 and 1.82.
The smoothing therefore depended on the gradient through the action curvature and not on shrinking the gradient change alone, in agreement with the anisotropy result of \cref{sec:exp_diag}.
The spatial displacement lowered the score by 1.4 to 1.6 times: with the temporal displacement alone, the score was 0.279, 0.333 and 0.517 against 0.205, 0.204 and 0.358.
Against the squared shift $\kappa^2\|\gs\|^2$, the curvature-weighted energy gave higher scores on all three tasks and a return that was higher on Walker, 4503 against 2583, and lower on Hopper, 2672 against 2783.
The \npr{} row used this validation-fixed weight without per-task selection, and its score was 5.1 to 13.0 times below that of Base.

\begin{table}[t]
\centering
\setlength{\tabcolsep}{4.6pt}
\caption{Validation of the greedy shift on the TD3 critics of \cref{tab:main}. The greedy motion $\Delta a^*$ is located by ascent on the critic. (a) uses all visited test states and the saddle-free shift $|B|^{-1}u$ at the executed action. (b) and (c) use $B^{-1}u$ at the greedy action and report medians over the states of (d): positive-definite $B$, an interior maximum and a matched maximizer branch. --: fewer than ten states.}
\label{tab:main_prop1}
\scriptsize
\begin{tabular}{@{}lcccccc@{}}
\toprule
Critic & Lunar & Pendulum & Reacher & Hopper & Walker & Ant\\
\midrule
\multicolumn{7}{@{}l}{\textit{(a) Rank correlation of the saddle-free shift with the greedy motion $\uparrow$}}\\
Base & 0.34 & 0.37 & \textbf{1.00} & 0.31 & 0.35 & 0.45\\
PAVE & 0.48 & 0.83 & \textbf{1.00} & 0.66 & 0.41 & 0.58\\
\rowcolor{blue!8}\npr{} (ours) & \textbf{0.97} & \textbf{0.90} & \textbf{1.00} & \textbf{0.79} & \textbf{0.98} & \textbf{0.85}\\
\midrule
\multicolumn{7}{@{}l}{\textit{(b) Cosine between $\gs$ and the greedy motion $\uparrow$}}\\
Base & \textbf{1.00} & 0.80 & \textbf{1.00} & 0.97 & -- & 0.83\\
PAVE & \textbf{1.00} & \textbf{0.94} & \textbf{1.00} & 0.99 & -- & 0.88\\
\rowcolor{blue!8}\npr{} (ours) & \textbf{1.00} & 0.74 & \textbf{1.00} & \textbf{1.00} & \textbf{1.00} & \textbf{0.99}\\
\midrule
\multicolumn{7}{@{}l}{\textit{(c) Relative error of $\gs$ (\%) $\downarrow$}}\\
Base & 13 & 53 & 0.5 & 50 & -- & 59\\
PAVE & 7.9 & \textbf{26} & 1.1 & 26 & -- & 61\\
\rowcolor{blue!8}\npr{} (ours) & \textbf{2.3} & 29 & \textbf{0.4} & \textbf{17} & \textbf{11} & \textbf{19}\\
\midrule
\multicolumn{7}{@{}l}{\textit{(d) States that satisfy the conditions of (b, c) (\%)}}\\
Base & 9 & 25 & 81 & 1 & 0 & 2\\
PAVE & 19 & 45 & 47 & 7 & 0 & 7\\
\rowcolor{blue!8}\npr{} (ours) & \textbf{45} & \textbf{50} & \textbf{90} & \textbf{10} & \textbf{11} & \textbf{23}\\
\bottomrule
\end{tabular}
\end{table}

\begin{table}[t]
\centering
\setlength{\tabcolsep}{4.6pt}
\caption{Validation of the hold-regret on the TD3 critics of \cref{tab:main}, on the states with an interior maximum. The realized value loss is $Q(s',a^*(s'))-Q(s',\ga)$. --: fewer than ten tested states; in (c), Pendulum is at the floating-point floor.}
\label{tab:main_regret}
\scriptsize
\begin{tabular}{@{}lcccccc@{}}
\toprule
Critic & Lunar & Pendulum & Reacher & Hopper & Walker & Ant\\
\midrule
\multicolumn{7}{@{}l}{\textit{(a) Realized value loss over $\tfrac12u^\top B^{-1}u$ (ideal: 1)}}\\
Base & 1.03 & 1.77 & \textbf{1.00} & 0.98 & -- & 1.14\\
PAVE & \textbf{1.01} & 1.20 & 1.06 & 0.96 & -- & 1.02\\
\rowcolor{blue!8}\npr{} (ours) & \textbf{1.01} & \textbf{1.13} & 1.01 & \textbf{0.99} & \textbf{1.01} & \textbf{0.99}\\
\midrule
\multicolumn{7}{@{}l}{\textit{(b) Rank correlation between the two $\uparrow$}}\\
Base & 0.99 & 0.93 & 0.97 & \textbf{0.99} & -- & 0.74\\
PAVE & \textbf{1.00} & \textbf{0.95} & 0.98 & \textbf{0.99} & -- & 0.87\\
\rowcolor{blue!8}\npr{} (ours) & \textbf{1.00} & 0.91 & \textbf{0.99} & \textbf{0.99} & \textbf{1.00} & \textbf{0.97}\\
\midrule
\multicolumn{7}{@{}l}{\textit{(c) Geometric-mean reduction of the remainder per halved transition (cubic: 8)}}\\
Base & 8.6 & -- & \textbf{8.0} & 5.4 & -- & 4.8\\
PAVE & 8.8 & -- & \textbf{8.0} & \textbf{8.0} & -- & 4.1\\
\rowcolor{blue!8}\npr{} (ours) & \textbf{8.1} & -- & 8.3 & 8.2 & \textbf{6.7} & \textbf{5.0}\\
\bottomrule
\end{tabular}
\end{table}

\subsection{Validation of greedy shift, hold-regret and anisotropy}
\label{sec:exp_diag}
\begin{figure}[t]
\centering
\includegraphics[width=\columnwidth]{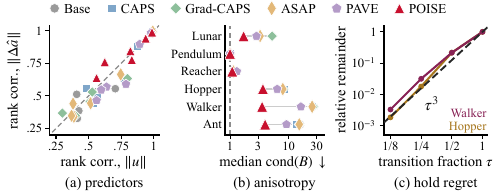}
\vspace{-1.6em}
\caption{Validation of the analysis with TD3 over 5 seeds. (a) Rank correlation of the executed action change with the saddle-free shift $\|\gs\|$ against that with $\|u\|$, one marker per task and method. (b) Median condition number of $B$; 1 is isotropic. (c) Hold-regret remainder on \npr{} critics against the cubic rate $\tau^3$ on Hopper and Walker.}
\label{fig:rank}\label{fig:aniso}\label{fig:regret}
\end{figure}

\noindent\textbf{Greedy shift.}
\Cref{tab:main_prop1} validated whether the greedy shift of \cref{def:gs} predicts the motion of the greedy action.
At each visited test state, we located the greedy actions before and after the transition by ascent on the critic and compared their difference with the shift.
On all visited states, the saddle-free shift of \cref{sec:method_metric} at the executed action ranked the size of the greedy motion with a correlation of 0.79 to 1.00 on the \npr{} critic, against 0.31 to 0.45 on the Base critic on five tasks.
The direct test of \cref{eq:increment} requires a positive-definite $B$, an interior maximum and a matched maximizer branch across the transition.
These conditions held together on 10\% to 90\% of the states of the \npr{} critic and on at most 25\% for Base on five tasks, although $B$ alone was positive definite on at least 91\% of the states for \npr{}.
On the states that satisfied all of these conditions, the shift matched the greedy motion as a vector: its cosine was 0.99 or more and its relative error 0.4\% to 19\% on five tasks, with Pendulum the exception at 0.74 and 29\%.
The link to the executed action was weaker: across the 150 critics of Base and the four baselines, $\|\gs\|$ ranked the executed action change below $\|u\|$ (\cref{fig:rank}a), with a mean rank correlation of 0.60 against 0.64.

\begin{figure*}[t]
\centering
\includegraphics[width=\textwidth]{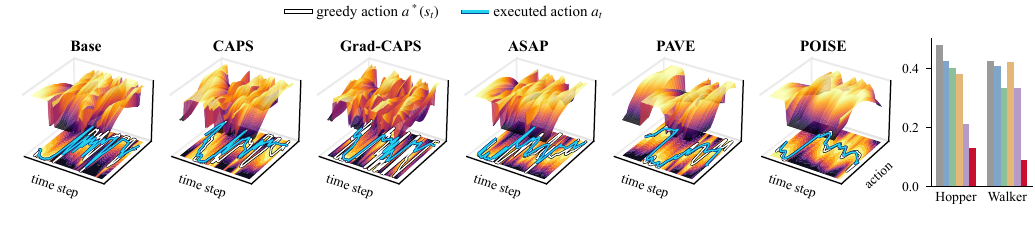}\\[-7pt]
{\footnotesize\makebox[\textwidth][l]{\rlap{\hspace{0.43\textwidth}\makebox[0pt]{(a) Critic landscape $Q(s_t,\cdot)$ along the principal action direction, Hopper test trajectory}}\rlap{\hspace{0.945\textwidth}\makebox[0pt]{(b) Greedy motion per step}}}}
\vspace{-1.2em}
\caption{Qualitative comparison of the critic landscapes and the motion of their greedy action (TD3). The bars of (b) follow the method order of (a).}
\label{fig:diag}
\end{figure*}

\noindent\textbf{Hold-regret.}
\label{sec:exp_regret}
\Cref{tab:main_regret} validated whether the hold-regret of \cref{cor:regret} matched the critic value lost by holding the previous greedy action.
On the critics of our method, the ratio of the realized loss to $\tfrac12u^\top B^{-1}u$ stayed between 0.99 and 1.13 and their rank correlation between 0.91 and 1.00 on all six tasks.
The relation is a property of any locally concave critic, and it held on the Base and PAVE critics as well, with the largest deviation a ratio of 1.77 for Base on Pendulum; Walker offered fewer than ten states with an interior maximum for both.
The remainder followed the cubic rate of \cref{cor:regret} on three tasks: over three halvings of the transition, its geometric-mean reduction per halving was 8.1 to 8.3 on Lunar, Reacher and Hopper, against the predicted 8, and 6.7 on Walker and 5.0 on Ant (\cref{tab:main_regret}c, \cref{fig:regret}c).

\begin{figure}[t]
\centering
\begin{minipage}[t]{0.475\columnwidth}
\centering
\includegraphics[width=\linewidth]{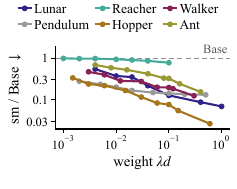}
\vspace{-1.6em}
\caption{sm against the weight $\lambda d$ over 5 or 10 runs.}
\label{fig:weight}
\end{minipage}\hfill
\begin{minipage}[t]{0.495\columnwidth}
\centering
\includegraphics[width=\linewidth]{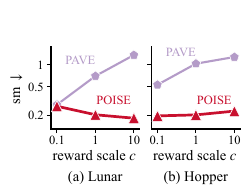}
\vspace{-1.6em}
\caption{sm against the reward scale $c$ over 5 seeds.}
\label{fig:scale}
\end{minipage}
\end{figure}
\begin{figure}[t]
\centering
\includegraphics[width=\columnwidth]{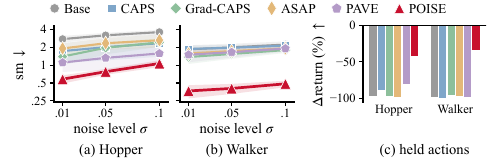}
\vspace{-1.6em}
\caption{sm under observation noise (a, b; one-std bands, 5 seeds) and return change under a four-step action hold (c).}
\label{fig:robust}
\end{figure}

\noindent\textbf{Anisotropy.}
\label{sec:exp_aniso}
\Cref{prop:gap} states that gradient changes of equal norm produce greedy shifts that differ by a factor of up to $\mathrm{cond}(B)$, and \cref{fig:aniso}b reported $\mathrm{cond}(B)$ of the learned critics.
The median condition number of the Base critic was 8.1 on Hopper, 26.8 on Walker and 15.7 on Ant. Gradient changes of equal norm received equal Euclidean penalties, whereas where $B\succ0$ \cref{prop:gap} lets their greedy shifts differ by up to this factor.
PAVE lowered the three values to 6.4, 16.2 and 9.2, and \npr{} to 3.6, 3.5 and 4.0.
The critics of Pendulum and Reacher were close to isotropic for every method, with condition numbers of at most 1.3.

\subsection{Critic geometry}
\label{sec:exp_geom}
\Cref{fig:diag} compared the critics that the methods produced.
At each step of a test trajectory, we evaluated $Q(s_t,\cdot)$ along the principal direction of the executed actions and marked its maximum, the greedy action on this slice, together with the executed action.
The Base critic had several ridges, its greedy action jumped between them from one step to the next, and the actor-side methods left this landscape nearly unchanged.
PAVE reduced the per-step motion of the greedy action on Hopper from 0.48 to 0.21, and the \npr{} critic kept a single ridge on this slice with a motion of 0.13, 3.7 times below Base on Hopper and 4.7 times on Walker.
The executed action followed the greedy action on this single ridge: their correlation was 0.97 on Hopper and 0.99 on Walker for our method, against 0.62 and 0.87 for Base.

\subsection{Robustness analysis}
\label{sec:exp_robust}
\label{sec:exp_scale}\label{sec:exp_noise}
The score of \npr{} fell as the weight grew on all six tasks (\cref{fig:weight}).
At the shared weight with rewards scaled by 0.1 and 10, the score of \npr{} stayed within 0.18 to 0.27 on Lunar, whereas that of PAVE rose from 0.28 to 1.37 (\cref{fig:scale}a); the Lunar return of \npr{} was 183, 226 and 240 at the scales 0.1, 1 and 10.
Under observation noise absent from training, \npr{} kept the lowest score at every noise level on Hopper and Walker (\cref{fig:robust}a, b).
With each action held for four steps, every baseline lost 81\% to 100\% of its return on both tasks, against 42\% and 34\% for \npr{} (\cref{fig:robust}c).
Under a one-step observation delay, the return of \npr{} changed by $+2\%$ on Hopper and $-24\%$ on Walker, against losses of 31\% to 57\% and 11\% to 43\% for the baselines.

\section{Conclusion}
\label{sec:conclusion}
We analyzed how the geometry of the critic moves its greedy action across a state transition and what critic value this motion costs.
The greedy shift measures this motion to first order, and its curvature-weighted energy measures the value lost by holding the previous greedy action.
\npr{} penalizes this energy on the critic with a saddle-free curvature metric and leaves the actor unchanged.
On six continuous control tasks, \npr{} reduced the TD3 smoothness score by $5.9\times$ on average, against $2.2\times$ for PAVE, and a single shared weight kept the lowest score among the regularized methods on all six tasks.
\paragraph{Limitations.}
This work evaluated \npr{} in simulation on state-based tasks with at most eight action dimensions.
Future work will address hardware deployment, image observations, on-policy methods, and curvature approximations for larger action spaces.

\section*{LLM Usage Statement}
We used Claude for coding assistance, language editing, literature search, and consistency checks, and ChatGPT for manuscript restructuring, language editing, and consistency checks. Claude assisted with implementing and refactoring experiment, measurement and figure code and with running the experiments under the authors' direction, which the authors read and verified. The reported experimental measurements originate from experiment outputs. The authors are responsible for the research design, experiments, final claims, citations, numerical results, and implementation descriptions.

{\small
\bibliographystyle{ieeenat_fullname}
\bibliography{refs}
}
\clearpage
\onecolumn
\begin{center}
  {\Large\textbf{\thetitle}}\\[0.5em]
  {\Large Supplementary Material}
\end{center}
\vspace{1em}
\suppressfloats[t]
\setcounter{section}{0}
\setcounter{figure}{0}
\setcounter{table}{0}
\setcounter{equation}{0}
\renewcommand{\thesection}{\Alph{section}}
\renewcommand{\thetable}{S\arabic{table}}
\renewcommand{\thefigure}{S\arabic{figure}}
\setcounter{algorithm}{0}
\renewcommand{\thealgorithm}{S\arabic{algorithm}}
\setcounter{proposition}{0}\renewcommand{\theproposition}{S\arabic{proposition}}
\setcounter{lemma}{0}\renewcommand{\thelemma}{S\arabic{lemma}}
\gdef\theHsection{supp.\arabic{section}}
\gdef\theHtable{supp.\arabic{table}}
\gdef\theHfigure{supp.\arabic{figure}}
\gdef\theHequation{supp.\arabic{equation}}
\gdef\theHalgorithm{supp.\arabic{algorithm}}
\gdef\theHproposition{supp.\arabic{proposition}}
\gdef\theHlemma{supp.\arabic{lemma}}
\gdef\theHsubsection{supp.\arabic{section}.\arabic{subsection}}
This supplement gives the protocol and the weight selection in \cref{supp:protocol}, the sensitivity to the weight in \cref{supp:weight}, the validity of the assumptions in \cref{supp:assump}, other regularizers in \cref{supp:mech}, the full ablation and robustness tables in \cref{sec:noise_supp}, the shared-weight results in \cref{supp:shared}, the action increments in \cref{supp:rmsda}, the cost in \cref{supp:curves}, weight sweeps of the baselines and the temporal-only form in \cref{supp:sweeps}, five further training seeds in \cref{supp:seeds}, SAC with automatic entropy tuning in \cref{supp:sacauto} and the auxiliary properties in \cref{supp:proofs}.

\section{Protocol, Hyperparameters and Statistics}
\label{supp:protocol}
\subsection{Algorithm}
\label{supp:alg}
\Cref{alg:poise} lists one critic update of \npr{} for TD3. SAC applies the same update to both critic heads and sums the penalty. The SAC TD loss had a factor of $\tfrac12$, which doubled the relative penalty weight per regularized critic head. Every SAC method followed the protocol of PAVE~\cite{lee2026stabilizing}: a batch of 100 transitions, 50 gradient updates after every 50 environment steps and an entropy coefficient fixed at $0.2$.
\begin{algorithm}[htbp]
\caption{One critic update of \npr{} (TD3). $\E$ is the batch mean and $\mathrm{sym}(M)=(M+M^\top)/2$. The penalty uses $Q_1$ alone, so $Q_2$ is trained on $\mathcal L_{\mathrm{TD}}$. The actor update is unchanged.}
\label{alg:poise}
\small
\begin{algorithmic}[1]
\Require batch $(s,a,r,s',\mathrm{done})$, critics $Q_{1,2}$ and targets $Q'_{1,2}$, target actor $\pi'$, weight $\lambda$, floor $\eta=0.1$
\State $y \gets r+\gamma(1-\mathrm{done})\min_j Q'_j\big(s',\pi'(s')+\xi\big)$ \Comment{TD target}
\State $\mathcal L_{\mathrm{TD}} \gets \sum_j \E\big(Q_j(s,a)-y\big)^2$
\State $g \gets \nabla_aQ_1(s,a)$, \ $u_{\mathrm T}\gets\nabla_aQ_1(s',a)-g$ \Comment{keep graph}
\State $\varepsilon\sim\{\pm1\}^{k}$, \ $u_{\mathrm S}\gets\nabla_aQ_1\big(s+\varepsilon\odot(s'-s),a\big)-g$
\State $B \gets -\mathrm{sym}(\nabla_a g)$ \Comment{$d$ backward passes}
\State $B=V\mathrm{diag}(\lambda_i)V^\top$, \ $\kappa\gets\mathrm{sg}\big(\E\,\tfrac1d\textstyle\sum_i|\lambda_i|\big)$ \Comment{batch scalar; per-sample means clamped at $10^{-8}$}
\State $|B| \gets B+\mathrm{sg}\big(V\mathrm{diag}(\max(|\lambda_i|,\eta\kappa)-\lambda_i)V^\top\big)$
\State $\mathcal L_{\mathrm{POISE}} \gets \kappa\,\E\big[u_{\mathrm T}^\top|B|^{-1}u_{\mathrm T}+u_{\mathrm S}^\top|B|^{-1}u_{\mathrm S}\big]$ \Comment{$d{\times}d$ solves}
\State update $Q_{1,2}$ on $\mathcal L_{\mathrm{TD}}+\lambda\,\mathcal L_{\mathrm{POISE}}$ with the base optimizer
\end{algorithmic}
\end{algorithm}

\subsection{Hyperparameters and weight selection}
\label{supp:results}
\Cref{tab:hp} lists the method-specific hyperparameters and \cref{tab:cost} the measured time per training update. Baseline values are the published per-task values of PAVE~\cite{lee2026stabilizing} and were not re-tuned. With SAC, CAPS and Grad-CAPS kept the listed values, and ASAP and PAVE used the published SAC values. Every method was trained for $10^5$ steps on Pendulum, $5\times10^5$ on Lunar and Reacher and $10^6$ on Hopper, Walker and Ant.
The weight of \npr{} was set in two stages. First, validation runs gave the shared rule $\lambda=\tilde\lambda/d$ with $\tilde\lambda=0.1$. Their five training seeds and their reset seeds were both disjoint from those of \cref{tab:main}, and Reacher had no validation runs. The runs behind this rule used $\lambda\in\{0.01,0.05\}$ and also 0.2 on Pendulum. The rule approximates the weights that these runs preferred, 0.05 at $d=2$ and 0.01 at $d=6$. Second, the weight in \cref{tab:main} was chosen per task among \npr{} runs on the five training seeds and the test reset seeds of \cref{tab:main}, as PAVE chose its weights. The candidates were the shared weight and one to five further weights per task with TD3, and the shared weight and one to three further weights with SAC. The choice took a low mean sm among weights whose paired return difference to Base had a 95\% bootstrap interval not entirely below zero. No weight qualified on TD3 Hopper and SAC Pendulum, where the weight of highest return was taken. The per-task weights were therefore selected on the reported seeds. \Cref{supp:shared} reports the shared weight without this selection, and \cref{fig:lambda} shows every trained weight.
\begin{table}[htbp]
\centering
\setlength{\tabcolsep}{2.5pt}
\caption{Method-specific hyperparameters (TD3 unless stated). CAPS: $(\lambda_T,\lambda_S,\sigma)$; Grad-CAPS: $\lambda_T$; ASAP: $(\lambda_T,\lambda_S,\lambda_P)$; PAVE: $(\lambda_1,\lambda_2,\lambda_3)$ with $\sigma{=}0.01$, $\delta{=}1$ fixed on all tasks; \npr{}: the weight $\lambda$ of $\mathcal L_{\mathrm{TD}}+\lambda\mathcal L_{\mathrm{POISE}}$ (not divided by $d$ again) with $\eta{=}0.1$; $^\dagger$: shared constant, $\lambda=\tilde\lambda/d$ with $\tilde\lambda=0.1$.}
\label{tab:hp}
\footnotesize
\begin{tabular}{lcccccc}
\toprule
Method & Lunar & Pendulum & Reacher & Hopper & Walker & Ant\\
action dimension $d$ & 2 & 1 & 2 & 3 & 6 & 8\\
\midrule
CAPS & .1, .5, .2 & 1, 5, .2 & .1, .5, .2 & .1, .5, .2 & .1, .5, .2 & .1, .5, .2\\
Grad-CAPS & 1 & 1 & 1 & 1 & 1 & 1\\
ASAP & .005, .03, 2 & .005, .03, 2 & .1, .1, 2 & .07, .3, 2 & .05, .3, 2 & .05, .3, 2\\
PAVE & .1, .1, .01 & 2, .005, 2 & .1, .1, .01 & .1, .005, .5 & .1, .1, .01 & .1, .005, .5\\
\npr{} (\cref{tab:main}) & .5 & .1$^\dagger$ & .02 & .002 & .017$^\dagger$ & .01\\
\npr{} (SAC, \cref{tab:main}) & .05$^\dagger$ & .005 & .05$^\dagger$ & .033$^\dagger$ & .01 & .0125$^\dagger$\\
\npr{} (shared, \cref{supp:shared}) & \multicolumn{6}{c}{$\tilde\lambda=0.1$, $\lambda=\tilde\lambda/d$}\\
\bottomrule
\end{tabular}
\end{table}

\begin{table}[htbp]
\centering
\setlength{\tabcolsep}{4pt}
\caption{Method hyperparameters listed in \cref{tab:hp} / those whose value varies across the six tasks and time per training update in ms, with the multiple of Base in parentheses (one RTX A5000, batch 256, mean over 2000 updates after 200 warm-up updates).}
\label{tab:cost}
\footnotesize
\resizebox{\textwidth}{!}{\begin{tabular}{llcccccc}
\toprule
Method & Listed / varying & Lunar ($d{=}2$) & Pendulum ($d{=}1$) & Reacher ($d{=}2$) & Hopper ($d{=}3$) & Walker ($d{=}6$) & Ant ($d{=}8$)\\
\midrule
Base & 0 / 0 & 3.7 (1.0$\times$) & 3.8 (1.0$\times$) & 4.0 (1.0$\times$) & 4.0 (1.0$\times$) & 3.9 (1.0$\times$) & 4.1 (1.0$\times$)\\
CAPS & 3 / 2 ($\lambda_T,\lambda_S,\sigma$) & 4.6 (1.2$\times$) & 4.8 (1.3$\times$) & 5.0 (1.2$\times$) & 5.0 (1.3$\times$) & 4.9 (1.3$\times$) & 5.2 (1.3$\times$)\\
Grad-CAPS & 1 / 0 ($\lambda_T$) & 4.6 (1.2$\times$) & 4.8 (1.3$\times$) & 5.0 (1.2$\times$) & 5.4 (1.4$\times$) & 5.0 (1.3$\times$) & 5.2 (1.3$\times$)\\
ASAP & 3 / 2 ($\lambda_T,\lambda_S,\lambda_P$) & 5.7 (1.5$\times$) & 6.0 (1.6$\times$) & 6.2 (1.5$\times$) & 6.7 (1.7$\times$) & 6.1 (1.6$\times$) & 6.4 (1.6$\times$)\\
PAVE & 5 / 3 ($\lambda_{1,2,3}$; $\sigma,\delta$ fixed) & 13.5 (3.6$\times$) & 14.4 (3.8$\times$) & 15.0 (3.7$\times$) & 15.3 (3.8$\times$) & 14.6 (3.7$\times$) & 15.0 (3.6$\times$)\\
\npr{} & 2 / 1 ($\lambda$; $\eta$ fixed) & 12.7 (3.4$\times$) & 11.2 (2.9$\times$) & 12.8 (3.2$\times$) & 14.4 (3.6$\times$) & 20.0 (5.1$\times$) & 23.5 (5.7$\times$)\\
\bottomrule
\end{tabular}}
\end{table}

\subsection{Statistics of the main table}
\label{supp:stats}
\Cref{tab:stats} lists the bootstrap 95\% interval of the mean smoothness factor of each method against Base and the tasks on which the paired return difference to Base excludes zero. Differences were paired by seed, and no correction for multiple comparisons was applied.
\begin{table}[t]
\centering
\setlength{\tabcolsep}{4pt}
\caption{Seed-level statistics of \cref{tab:main} (Gym/MuJoCo, 5 training seeds, shared reset seeds). Factor: mean over the six tasks of Base sm / method sm, with a 95\% bootstrap interval over seeds (10k resamples). re$\downarrow$ / re$\uparrow$: tasks whose paired per-seed return difference against Base has a 95\% bootstrap interval entirely below / above zero; Clear sm wins: tasks where the method has the lowest mean sm and its paired difference against the second-lowest method has an interval excluding zero.}
\label{tab:stats}
\footnotesize
\begin{tabular}{llcp{0.17\linewidth}p{0.17\linewidth}p{0.2\linewidth}}
\toprule
Backbone & Method & sm factor [95\% CI] & re$\downarrow$ & re$\uparrow$ & clear sm wins\\
\midrule
TD3 & CAPS & 1.9$\times$ [1.6, 2.3] & 2 (Pendulum, Hopper) & 0 & 0\\
TD3 & Grad-CAPS & 1.6$\times$ [1.4, 1.8] & 0 & 2 (Walker, Ant) & 1 (Reacher)\\
TD3 & ASAP & 1.1$\times$ [0.9, 1.3] & 1 (Hopper) & 0 & 0\\
TD3 & PAVE & 2.2$\times$ [1.8, 2.5] & 1 (Reacher) & 3 (Pendulum, Walker, Ant) & 0\\
TD3 & \npr{} & 5.9$\times$ [5.2, 6.6] & 1 (Hopper) & 1 (Ant) & 5 (Lunar, Pendulum, Hopper, Walker, Ant)\\
\midrule
SAC & CAPS & 1.2$\times$ [1.1, 1.3] & 1 (Pendulum) & 0 & 0\\
SAC & Grad-CAPS & 1.3$\times$ [1.2, 1.3] & 1 (Reacher) & 0 & 0\\
SAC & ASAP & 1.4$\times$ [1.4, 1.5] & 1 (Reacher) & 0 & 0\\
SAC & PAVE & 1.3$\times$ [1.2, 1.5] & 2 (Pendulum, Walker) & 0 & 0\\
SAC & \npr{} & 2.3$\times$ [2.2, 2.5] & 1 (Pendulum) & 0 & 4 (Pendulum, Hopper, Walker, Ant)\\
\bottomrule
\end{tabular}
\end{table}

\section{Sensitivity to the Weight and the Curvature Floor}
\label{supp:weight}
\begin{figure*}[htbp]
\centering
\includegraphics[width=\textwidth]{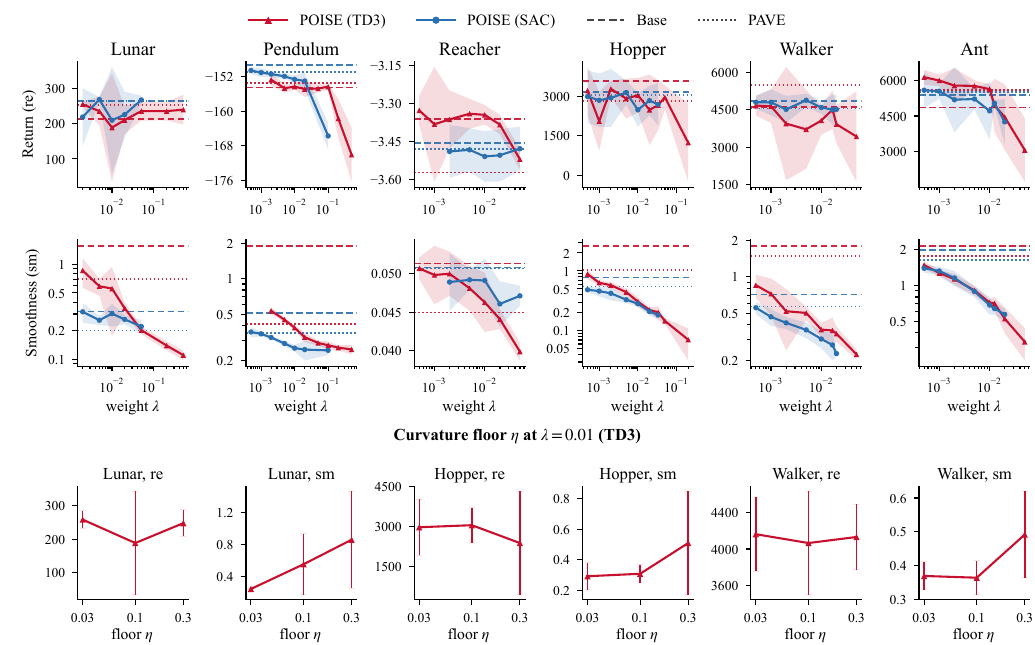}
\caption{Sensitivity of \npr{} to its weight $\lambda$ (top two rows) and floor $\eta$ (bottom row), test resets, 95\% $t$-intervals over seeds. Every trained weight is shown; runs on the seeds of \cref{tab:main} and sweep runs on other training seeds are pooled at equal $\lambda$ (5 or 10 seeds per point). Shared-constant runs are placed at $\lambda=0.1/d$. Dashed: Base; dotted: PAVE, in the color of the backbone (red TD3, blue SAC). Bottom row: TD3, $\lambda=0.01$, where $\eta=0.1$ is the default.}
\label{fig:lambda}
\end{figure*}

\Cref{fig:lambda} shows \npr{} on all six tasks over every trained weight, on the test reset seeds. The return fell beyond a task-dependent weight; this weight ranged from $0.01$ on Ant to beyond $0.5$ on Lunar, so \cref{prop:scale} removes the dependence of $\lambda$ on reward units, not on the task. The floors $\eta=0.03$ and $0.1$ gave scores within 0.02 of each other on Hopper and Walker, and $\eta=0.3$ gave the highest score on all three tested tasks.

\section{Assumption Validity}
\label{supp:assump}
\Cref{tab:assump} measures how widely trained TD3 critics satisfy \cref{ass:A,ass:B}: the rate of states where $B$ is not positive definite, the saturation rate, the gap between the actor and the greedy action, and rank correlations with the executed action change. The greedy action was located by Newton ascent on the critic, and the applicable states of \cref{tab:main_prop1} are those with positive definite $B$, an interior maximum, a converged ascent and no jump to a different maximum.
\begin{table}[htbp]
\centering
\setlength{\tabcolsep}{5pt}
\caption{Assumption checks on trained critics (TD3, 5 seeds, test trajectories; mean over seeds of per-critic medians). Each quantity is given for Base, PAVE, \npr{} at the per-task weight of \cref{tab:main}, and \npr{} at the shared weight $\tilde\lambda=0.1$ ($\lambda=\tilde\lambda/d$), which equals the weight of \cref{tab:main} on Pendulum and Walker. $\gs$ uses the saddle-free metric of \npr{}; $a^*$ is found by Newton ascent on the critic from the executed action.}
\label{tab:assump}
\footnotesize
\begin{tabular}{lcccccc}
\toprule
Quantity / critic & Lunar & Pendulum & Reacher & Hopper & Walker & Ant\\
\midrule
\multicolumn{7}{l}{\textit{non-PD rate of $B$}}\\
\quad Base & 0.60 & 0.31 & 0.00 & 0.75 & 0.82 & 0.82\\
\quad PAVE & 0.25 & 0.00 & 0.00 & 0.20 & 0.63 & 0.32\\
\quad \npr{}, weight of \cref{tab:main} & 0.00 & 0.00 & 0.00 & 0.08 & 0.00 & 0.00\\
\quad \npr{}, shared $\tilde\lambda=0.1$ & 0.02 & 0.00 & 0.00 & 0.00 & 0.00 & 0.00\\
\midrule
\multicolumn{7}{l}{\textit{saturated actions ($|a|>0.95$)}}\\
\quad Base & 0.26 & 0.32 & 0.00 & 0.24 & 0.31 & 0.03\\
\quad PAVE & 0.16 & 0.03 & 0.00 & 0.13 & 0.37 & 0.03\\
\quad \npr{}, weight of \cref{tab:main} & 0.01 & 0.02 & 0.00 & 0.07 & 0.15 & 0.00\\
\quad \npr{}, shared $\tilde\lambda=0.1$ & 0.02 & 0.02 & 0.00 & 0.05 & 0.15 & 0.00\\
\midrule
\multicolumn{7}{l}{\textit{actor gap $\|a_t-a^*(s_t)\|$}}\\
\quad Base & 0.73 & 0.30 & 0.01 & 1.07 & 1.25 & 1.22\\
\quad PAVE & 0.42 & 0.24 & 0.02 & 0.37 & 1.16 & 0.48\\
\quad \npr{}, weight of \cref{tab:main} & 0.10 & 0.18 & 0.01 & 0.19 & 0.14 & 0.09\\
\quad \npr{}, shared $\tilde\lambda=0.1$ & 0.17 & 0.18 & 0.01 & 0.06 & 0.14 & 0.09\\
\midrule
\multicolumn{7}{l}{\textit{$\mathrm{rc}(\|\gs\|,\|\Delta a\|)$}}\\
\quad Base & 0.51 & 0.33 & 0.99 & 0.39 & 0.50 & 0.56\\
\quad PAVE & 0.57 & 0.90 & 0.98 & 0.55 & 0.46 & 0.59\\
\quad \npr{}, weight of \cref{tab:main} & 0.76 & 0.94 & 0.99 & 0.64 & 0.84 & 0.78\\
\quad \npr{}, shared $\tilde\lambda=0.1$ & 0.68 & 0.94 & 0.98 & 0.81 & 0.84 & 0.76\\
\midrule
\multicolumn{7}{l}{\textit{$\mathrm{rc}(\|u\|,\|\Delta a\|)$}}\\
\quad Base & 0.41 & 0.41 & 0.99 & 0.42 & 0.56 & 0.70\\
\quad PAVE & 0.64 & 0.89 & 0.98 & 0.60 & 0.56 & 0.79\\
\quad \npr{}, weight of \cref{tab:main} & 0.62 & 0.94 & 0.99 & 0.56 & 0.82 & 0.83\\
\quad \npr{}, shared $\tilde\lambda=0.1$ & 0.56 & 0.94 & 0.98 & 0.72 & 0.82 & 0.70\\
\midrule
\multicolumn{7}{l}{\textit{$\mathrm{rc}(\|\gs\|,\|a^*(s')-a^*(s)\|)$}}\\
\quad Base & 0.34 & 0.37 & 1.00 & 0.31 & 0.35 & 0.45\\
\quad PAVE & 0.48 & 0.83 & 1.00 & 0.66 & 0.41 & 0.58\\
\quad \npr{}, weight of \cref{tab:main} & 0.97 & 0.90 & 1.00 & 0.79 & 0.98 & 0.85\\
\quad \npr{}, shared $\tilde\lambda=0.1$ & 0.89 & 0.90 & 1.00 & 0.98 & 0.98 & 0.83\\
\midrule
\multicolumn{7}{l}{\textit{$\mathrm{cond}(B)$, median}}\\
\quad Base & 3.0 & 1.0 & 1.1 & 8.1 & 26.8 & 15.7\\
\quad PAVE & 2.8 & 1.0 & 1.3 & 6.4 & 16.2 & 9.2\\
\quad \npr{}, weight of \cref{tab:main} & 1.7 & 1.0 & 1.1 & 3.6 & 3.5 & 4.0\\
\quad \npr{}, shared $\tilde\lambda=0.1$ & 1.9 & 1.0 & 1.1 & 2.3 & 3.5 & 5.1\\
\bottomrule
\end{tabular}
\end{table}

\paragraph{Base and PAVE critics.}
On Lunar, Hopper, Walker and Ant, the non-PD rates were 0.60 to 0.82 for Base and 0.20 to 0.63 for PAVE. Over all 150 trained baseline critics the mean rank correlation with the executed action change was 0.60 for $\|\gs\|$ and 0.64 for $\|u\|$. \Cref{ass:A,ass:B} held more often on \npr{} critics, and the statements of the main paper apply to any critic that satisfies them.
\paragraph{\npr{} critics.}
At the weight of \cref{tab:main} the non-PD rate was 0.08 on Hopper and 0.00 on the other tasks, and $\|\gs\|$ was the better predictor on Lunar, Hopper and Walker, tied on Pendulum and Reacher and worse on Ant, with 0.78 against 0.83.\par The condition number of $B$ is $\lambda_{\max}/\lambda_{\min}$ at the states with $B\succ0$, capped at $10^6$. Its median was taken per seed and averaged over seeds. The cosine of \cref{tab:main_prop1} was computed at each applicable state with $10^{-12}$ added to the product of the two norms in its denominator, and the table averages the per-seed medians. At $d=1$ the cosine of two nonzero scalars is $\pm1$, so a median strictly between $-1$ and $1$ arises only where the product of the two magnitudes is of the order of $10^{-12}$. The Pendulum value of 0.74 is the mean of 1.00 on three seeds and of 0.53 and 0.15 on two seeds, whose median states had a greedy motion and a shift with a product near $10^{-12}$.
\paragraph{Remainder scaling.}
\Cref{tab:cor1tau} shortens the transition to $s+\tau\Delta s$ and lists the error of \cref{cor:regret}, with the critic evaluated in double precision. On Lunar and Reacher the median error of the \npr{} critics fell by a factor of 7.9 to 8.8 at each halving, the cubic rate. On Hopper, Walker and Ant the first halving gave 4.6 to 4.8 and the later halvings gave 4.0 to 12.2. The main paper reports the geometric mean over the three halvings, which was 8.2 on Hopper, 6.7 on Walker and 5.0 on Ant. On Pendulum the error stayed near $10^{-14}$ at every $\tau$.
\begin{table}[htbp]
\centering
\setlength{\tabcolsep}{4pt}
\caption{Scaling of the remainder of \cref{cor:regret} (TD3 critics of \cref{tab:main}, 5 seeds, critic evaluated in double precision). At the applicable states of \cref{tab:main_prop1} whose Newton ascent converged at every $\tau$ ($n$), the transition is shortened to $s_\tau=s+\tau\Delta s$ and the median absolute error $|R_Q-\tfrac12u_\tau^\top B^{-1}u_\tau|$ is measured, where $R_Q=Q(s_\tau,a^*(s_\tau))-Q(s_\tau,a^*(s))$; a cubic remainder predicts a factor of 8 per halving of $\tau$, a quadratic one a factor of 4. Decay: ratio of the median error between consecutive $\tau$.}
\label{tab:cor1tau}
\footnotesize
\begin{tabular}{lcccccc}
\toprule
Quantity / critic & Lunar & Pendulum & Reacher & Hopper & Walker & Ant\\
\midrule
\multicolumn{7}{l}{\textit{states $n$ (of 1000)}}\\
\quad Base & 87 & 253 & 813 & 10 & 2 & 19\\
\quad PAVE & 190 & 446 & 467 & 70 & 2 & 67\\
\quad \npr{}, weight of \cref{tab:main} & 451 & 504 & 900 & 101 & 108 & 221\\
\midrule
\multicolumn{7}{l}{\textit{median $|R_Q|$ at $\tau=1$}}\\
\quad Base & 0.00394 & 5.12e-13 & 3.66e-05 & 0.028 & 0.195 & 1.715\\
\quad PAVE & 0.00292 & 1.67e-09 & 1.6e-05 & 0.04 & 0.408 & 1.551\\
\quad \npr{}, weight of \cref{tab:main} & 0.000712 & 1.01e-09 & 4.65e-05 & 0.042 & 0.099 & 0.378\\
\midrule
\multicolumn{7}{l}{\textit{median error at $\tau=1$}}\\
\quad Base & 0.000202 & 2.08e-14 & 6.44e-08 & 0.00452 & 0.06 & 0.411\\
\quad PAVE & 0.000136 & 1.32e-13 & 3.68e-08 & 0.00448 & 0.064 & 0.446\\
\quad \npr{}, weight of \cref{tab:main} & 8.51e-06 & 5.3e-14 & 1.17e-07 & 0.00214 & 0.00461 & 0.034\\
\midrule
\multicolumn{7}{l}{\textit{decay $\tau=1\to0.5$}}\\
\quad Base & 9.73 & 0.99 & 7.94 & 4.09 & 4.31 & 1.86\\
\quad PAVE & 10.95 & 3.49 & 7.94 & 9.38 & 4.37 & 3.84\\
\quad \npr{}, weight of \cref{tab:main} & 8.26 & 1.4 & 8.84 & 4.61 & 4.65 & 4.78\\
\midrule
\multicolumn{7}{l}{\textit{decay $\tau=0.5\to0.25$}}\\
\quad Base & 7.39 & 1.01 & 7.92 & 5.93 & 11.84 & 7.3\\
\quad PAVE & 7.56 & 1.09 & 8.56 & 6.8 & 6.08 & 3.48\\
\quad \npr{}, weight of \cref{tab:main} & 7.93 & 1.12 & 8.01 & 12.2 & 6.85 & 4.03\\
\midrule
\multicolumn{7}{l}{\textit{decay $\tau=0.25\to0.125$}}\\
\quad Base & 8.73 & 1.04 & 8.18 & 6.55 & 10.97 & 8.18\\
\quad PAVE & 8.29 & 1.26 & 7.43 & 8.14 & 6.96 & 5.25\\
\quad \npr{}, weight of \cref{tab:main} & 8.15 & 1.12 & 8.02 & 9.68 & 9.6 & 6.54\\
\bottomrule
\end{tabular}
\end{table}

\paragraph{Replayed actions.}
The loss evaluates $\gs$ at replayed actions, while \cref{eq:increment} concerns the greedy action. If $u$ and $B$ are evaluated at an action $a$ with $\|a-\ga\|=\rho$ small enough that $B(s,a)\succ0$, then $\gs(a)-\gs(\ga)=O(\rho\|\Delta s\|)$. With the TD3 exploration noise added to the actor action on \npr{} critics, the median relative difference between $\gs$ at the two points was 0.05 to 0.22 on five tasks and 0.34 on Pendulum, and the rank correlation of $\|\gs\|$ with $\|\Delta a\|$ changed by at most 0.02. This test covered added exploration noise; replay staleness was unmeasured.
\paragraph{Landscape slice.}
In \cref{fig:diag} the direction $v$ is the first principal direction of the executed actions over the plotted steps of the first test episode. The slice at step $t$ passes through the executed action, $a(y)=\mathrm{clip}(a_t+(y-v^\top a_t)v,-1,1)$, with $y$ on a fixed grid of 121 points, and the greedy coordinate $y^*_t$ maximizes $Q_1(s_t,a(y))$ on this grid. The motion in panel (b) is the mean of $|y^*_{t+1}-y^*_t|$, averaged over seeds. It is a one-dimensional quantity on a slice that moves with $a_t$.

\section{Alternative Regularizers}
\label{supp:mech}
\paragraph{Relation to PAVE.}
PAVE~\cite{lee2026stabilizing} penalizes the change of the action gradient and the curvature in two separate terms. Its curvature term $\mathrm{relu}(v^\top\Haa v+\delta)$ raises the curvature along random probe directions $v$ until the fixed margin $\delta$ is met, whatever the size of the gradient change. \npr{} measures the gradient change in the inverse curvature metric, with partial derivatives given in \cref{supp:surrogate}. The Euclidean row of \cref{tab:ablation_full} has the form of the transition term of PAVE without its curvature term.
\subsection{LipsNet and LipsNet++}
\label{supp:lipsnet}
Both methods replace only the TD3 actor; the critic and every other setting are those of Base. The network $f$ of both methods has the hidden sizes and activation of the Base actor. LipsNet~\cite{song2023lipsnet} was run with $K$ initialized at 1, the value of the official example, and $\lambda\in\{10^{-5},10^{-3}\}$, because the initialization at 50 saturated the tanh output. LipsNet++~\cite{song2025lipsnet} was run with $\lambda_h=10^{-3}$ and $\lambda_k\in\{10^{-2},10^{-3}\}$; its actor received the last five observations and its critic the latest one. The weight of each method was chosen per task by the rule of \cref{supp:results}.
The rows ``+\,\npr{}'' of \cref{tab:lipsnet} add the loss of \npr{} to the critic at the weight of \cref{tab:main} without retuning. \npr{} lowered the sm of LipsNet by $1.2$--$4.3\times$ and that of LipsNet++ by $1.1$--$13\times$. The return rose on Lunar for both actors and fell on Hopper and Walker.
\begin{table*}[t]
\centering
\setlength{\tabcolsep}{3.0pt}
\renewcommand{\arraystretch}{0.95}
\caption{Combination with Lipschitz-constrained actors under TD3 (5 seeds, standard deviation in parentheses). LipsNet~\cite{song2023lipsnet} and LipsNet++~\cite{song2025lipsnet} alone and with \npr{} on the critic at its weight of \cref{tab:main}. \textbf{Bold}: best; \underline{underline}: second; Base is not ranked.}
\label{tab:lipsnet}
\scriptsize
\begin{tabular}{@{}lcccccccccccc@{}}
\toprule
& \multicolumn{2}{c}{Lunar} & \multicolumn{2}{c}{Pendulum} & \multicolumn{2}{c}{Reacher} & \multicolumn{2}{c}{Hopper} & \multicolumn{2}{c}{Walker} & \multicolumn{2}{c}{Ant}\\
\cmidrule(lr){2-3} \cmidrule(lr){4-5} \cmidrule(lr){6-7} \cmidrule(lr){8-9} \cmidrule(lr){10-11} \cmidrule(lr){12-13}
Method & re$\uparrow$ & sm$\downarrow$ & re$\uparrow$ & sm$\downarrow$ & re$\uparrow$ & sm$\downarrow$ & re$\uparrow$ & sm$\downarrow$ & re$\uparrow$ & sm$\downarrow$ & re$\uparrow$ & sm$\downarrow$\\
\midrule
TD3 Base & 214.0{\tiny\,(65.3)} & 1.580{\tiny\,(0.571)} & -154.5{\tiny\,(0.78)} & 1.900{\tiny\,(0.381)} & -3.36{\tiny\,(0.08)} & 0.051{\tiny\,(0.001)} & 3595{\tiny\,(58.3)} & 2.645{\tiny\,(0.363)} & 4596{\tiny\,(562.3)} & 1.818{\tiny\,(0.307)} & 4856{\tiny\,(1008)} & 2.158{\tiny\,(0.174)}\\
\multicolumn{13}{l}{\cellcolor{gray!12}\textit{Lipschitz-constrained actor}}\\
LipsNet~\cite{song2023lipsnet} & 219.5{\tiny\,(22.3)} & 0.427{\tiny\,(0.172)} & -156.5{\tiny\,(3.72)} & 0.871{\tiny\,(0.146)} & \textbf{-3.35}{\tiny\,(0.05)} & 0.052{\tiny\,(0.002)} & 3170{\tiny\,(770.7)} & 2.153{\tiny\,(0.302)} & \textbf{5167}{\tiny\,(648.0)} & 1.562{\tiny\,(0.170)} & 4661{\tiny\,(2131)} & 1.831{\tiny\,(0.689)}\\
LipsNet++~\cite{song2025lipsnet} & 144.2{\tiny\,(242.2)} & 1.072{\tiny\,(1.192)} & \textbf{-154.2}{\tiny\,(2.43)} & 2.242{\tiny\,(1.479)} & -3.38{\tiny\,(0.05)} & 0.049{\tiny\,(0.002)} & \textbf{3475}{\tiny\,(163.9)} & 2.625{\tiny\,(0.151)} & \underline{5142}{\tiny\,(407.1)} & 1.514{\tiny\,(0.218)} & 4814{\tiny\,(1067)} & 2.096{\tiny\,(0.154)}\\
\multicolumn{13}{l}{\cellcolor{gray!12}\textit{With \npr{} on the critic}}\\
\rowcolor{blue!8}LipsNet + \npr{} & \textbf{257.0}{\tiny\,(10.9)} & \underline{0.103}{\tiny\,(0.014)} & -158.5{\tiny\,(8.72)} & \textbf{0.263}{\tiny\,(0.034)} & -3.39{\tiny\,(0.08)} & 0.045{\tiny\,(0.002)} & 2885{\tiny\,(753.5)} & \textbf{0.503}{\tiny\,(0.115)} & 3804{\tiny\,(509.9)} & 0.368{\tiny\,(0.023)} & \underline{5388}{\tiny\,(169.0)} & \underline{0.694}{\tiny\,(0.079)}\\
\rowcolor{blue!8}LipsNet++ + \npr{} & 176.4{\tiny\,(81.3)} & \textbf{0.082}{\tiny\,(0.006)} & -156.9{\tiny\,(5.55)} & 0.287{\tiny\,(0.020)} & -3.53{\tiny\,(0.18)} & \textbf{0.043}{\tiny\,(0.003)} & 2674{\tiny\,(961.7)} & \underline{0.520}{\tiny\,(0.064)} & 3669{\tiny\,(1221)} & \textbf{0.281}{\tiny\,(0.038)} & 4716{\tiny\,(761.4)} & \textbf{0.589}{\tiny\,(0.124)}\\
\npr{}, Base actor & \underline{239.8}{\tiny\,(33.9)} & 0.110{\tiny\,(0.011)} & \underline{-154.4}{\tiny\,(0.55)} & \underline{0.272}{\tiny\,(0.017)} & \underline{-3.36}{\tiny\,(0.04)} & \underline{0.044}{\tiny\,(0.001)} & \underline{3208}{\tiny\,(623.1)} & 0.561{\tiny\,(0.101)} & 4503{\tiny\,(621.8)} & \underline{0.358}{\tiny\,(0.082)} & \textbf{5559}{\tiny\,(292.4)} & 0.726{\tiny\,(0.092)}\\
\bottomrule
\end{tabular}
\end{table*}

\subsection{Critic-side controls}
\label{supp:critic}
\Cref{tab:critic} regularizes the critic without the curvature metric: a penalty on $\|\nabla_aQ\|^2$ at replayed pairs (GP-critic) or spectral normalization of the penultimate layer of each head (SN-critic). The lowest sm of GP-critic was 1.29 on Lunar and 1.93 on Pendulum, against 0.11 and 0.27 for \npr{}, and SN-critic lowered sm by at most 36\%.
\begin{table}[htbp]
\centering
\setlength{\tabcolsep}{4pt}
\caption{Critic-side controls (TD3, test, mean over the 5 seeds of \cref{tab:main}). GP-critic: $\lambda\,\mathbb E\|\nabla_aQ_k\|^2$ on both critic heads; SN-critic: spectral normalization of the penultimate layer of each head~\cite{gogianu2021spectral}. Actor and TD3 update as in Base. \textbf{Bold}: lowest sm.}
\label{tab:critic}
\footnotesize
\begin{tabular}{lrrrrrrrrrrrr}
\toprule
& \multicolumn{2}{c}{Lunar} & \multicolumn{2}{c}{Pendulum} & \multicolumn{2}{c}{Reacher} & \multicolumn{2}{c}{Hopper} & \multicolumn{2}{c}{Walker} & \multicolumn{2}{c}{Ant}\\
\cmidrule(lr){2-3} \cmidrule(lr){4-5} \cmidrule(lr){6-7} \cmidrule(lr){8-9} \cmidrule(lr){10-11} \cmidrule(lr){12-13}
Method & re & sm & re & sm & re & sm & re & sm & re & sm & re & sm\\
\midrule
\color{gray}Base & \color{gray}214.0 & \color{gray}1.580 & \color{gray}-154.5 & \color{gray}1.900 & \color{gray}-3.36 & \color{gray}0.051 & \color{gray}3595 & \color{gray}2.645 & \color{gray}4596 & \color{gray}1.818 & \color{gray}4856 & \color{gray}2.158\\
GP-critic, $\lambda=10^{-4}$ & 176.0 & 2.155 & -155.2 & 2.319 & -3.33 & 0.050 & 3217 & 2.939 & 4491 & 1.855 & 5304 & 2.116\\
GP-critic, $\lambda=10^{-3}$ & 253.2 & 1.416 & -155.4 & 3.218 & -3.50 & 0.053 & 2671 & 2.218 & 4506 & 2.053 & 4835 & 2.142\\
GP-critic, $\lambda=10^{-2}$ & 258.4 & 1.292 & -155.7 & 1.927 & -3.45 & 0.062 & 2474 & 2.307 & 4975 & 1.981 & 1512 & 1.011\\
SN-critic & 205.8 & 1.531 & -156.3 & 1.221 & -3.59 & 0.049 & 3502 & 2.389 & 4460 & 1.459 & 5445 & 2.223\\
\npr{} & 239.8 & \textbf{0.110} & -154.4 & \textbf{0.272} & -3.36 & \textbf{0.044} & 3208 & \textbf{0.561} & 4503 & \textbf{0.358} & 5559 & \textbf{0.726}\\
\bottomrule
\end{tabular}
\end{table}

\section{Ablation and Robustness Tables}
\label{sec:noise_supp}
\Cref{tab:ablation_full} repeats \cref{tab:ablation} with standard deviations over seeds and adds the build-up from Base. The Euclidean penalty and the detached scalar metric did not lower sm below 1.66 on any task. The gradient through the scalar curvature lowered it to 0.38 to 0.82, and the anisotropic metric lowered it by a further 1.9 to 2.5 times. With the anisotropic metric detached the sm stayed at 1.90, 3.05 and 1.70, above that of \npr{} on all three tasks. \Cref{supp:surrogate} gives the partial derivatives of the penalty. The $\kappa^2\|\gs\|^2$ form was run at the same nominal weight as \npr{}, not at a matched regularization strength. It had a lower sm on all three tasks and a higher return on Hopper, and its Walker return was 2583 against 4503. The last two rows of \cref{tab:ablation_full} test the two stop-gradients of \npr{}. With the gradient through $\kappa$ kept, the sm rose from 0.205, 0.204 and 0.358 to 0.290, 0.472 and 0.604. With the gradient through the spectral correction kept, the sm rose to 1.897, 1.068 and 0.806 and the Lunar return fell from 226 to 142. Both stop-gradients therefore contributed to the smoothing.
\begin{table}[htbp]
\centering
\setlength{\tabcolsep}{1.8pt}
\caption{Component analysis with standard deviations (TD3, test, 5 seeds, std in parentheses; shared weight $\tilde\lambda=0.1$). Top: \npr{} (bold) built up from Base; bottom: one change to \npr{}. $\bar\lambda=\frac1d\sum_i|\lambda_i(B)|$, detached; the differentiable scalar metric uses the signed mean curvature $\mathrm{tr}(B)/d$ floored at $\eta\kappa$, which differs from $\bar\lambda$ where $B$ is not positive definite. The last two rows remove one stop-gradient of \npr{}: on the scale $\kappa$, or on the spectral correction of $|B|$ with the floor still detached.}
\label{tab:ablation_full}
\footnotesize
\begin{tabular}{@{}lcccccc@{}}
\toprule
& \multicolumn{2}{c}{Lunar} & \multicolumn{2}{c}{Hopper} & \multicolumn{2}{c}{Walker}\\
\cmidrule(lr){2-3}\cmidrule(lr){4-5}\cmidrule(lr){6-7}
Configuration & re$\uparrow$ & sm$\downarrow$ & re$\uparrow$ & sm$\downarrow$ & re$\uparrow$ & sm$\downarrow$\\
\midrule
\rowcolor{gray!10}\color{gray}Base & \color{gray}214.0{\tiny\,(65.3)} & \color{gray}1.580{\tiny\,(0.571)} & \color{gray}3595{\tiny\,(58.3)} & \color{gray}2.645{\tiny\,(0.363)} & \color{gray}4596{\tiny\,(562.3)} & \color{gray}1.818{\tiny\,(0.307)}\\
+ shift penalty, Euclidean & 170.3{\tiny\,(111.8)} & 2.479{\tiny\,(1.356)} & 2662{\tiny\,(925.4)} & 2.792{\tiny\,(0.867)} & 3860{\tiny\,(2095)} & 1.660{\tiny\,(0.499)}\\
+ scalar metric $\bar\lambda^{-1}I$ & 204.0{\tiny\,(61.2)} & 1.736{\tiny\,(0.807)} & 3384{\tiny\,(513.7)} & 3.504{\tiny\,(0.497)} & 4826{\tiny\,(812.2)} & 1.936{\tiny\,(0.381)}\\
+ gradient through curvature ($\mathrm{tr}(B)/d$, floored at $\eta\kappa$) & 250.7{\tiny\,(17.1)} & 0.380{\tiny\,(0.132)} & 2755{\tiny\,(915.2)} & 0.514{\tiny\,(0.083)} & 4010{\tiny\,(235.8)} & 0.815{\tiny\,(0.127)}\\
\textbf{+ anisotropic $|B|^{-1}$ (\npr{})} & \textbf{226.0}{\tiny\,(44.2)} & \textbf{0.205}{\tiny\,(0.028)} & \textbf{2672}{\tiny\,(766.5)} & \textbf{0.204}{\tiny\,(0.043)} & \textbf{4503}{\tiny\,(621.8)} & \textbf{0.358}{\tiny\,(0.082)}\\
\midrule
curvature detached & 227.5{\tiny\,(53.4)} & 1.902{\tiny\,(0.993)} & 2673{\tiny\,(966.5)} & 3.050{\tiny\,(0.613)} & 4982{\tiny\,(536.4)} & 1.699{\tiny\,(0.389)}\\
no spatial term & 219.6{\tiny\,(53.7)} & 0.279{\tiny\,(0.081)} & 2305{\tiny\,(1047)} & 0.333{\tiny\,(0.068)} & 4517{\tiny\,(391.2)} & 0.517{\tiny\,(0.128)}\\
$\kappa^2\|\gs\|^2$ form & 211.6{\tiny\,(30.5)} & 0.167{\tiny\,(0.015)} & 2783{\tiny\,(421.0)} & 0.122{\tiny\,(0.027)} & 2583{\tiny\,(1542)} & 0.217{\tiny\,(0.035)}\\
gradient through $\kappa$ kept & 260.5{\tiny\,(11.0)} & 0.290{\tiny\,(0.037)} & 2819{\tiny\,(893.4)} & 0.472{\tiny\,(0.048)} & 3823{\tiny\,(497.6)} & 0.604{\tiny\,(0.072)}\\
gradient through the correction kept & 142.0{\tiny\,(105.8)} & 1.897{\tiny\,(0.697)} & 3237{\tiny\,(684.2)} & 1.068{\tiny\,(0.519)} & 5065{\tiny\,(233.7)} & 0.806{\tiny\,(0.140)}\\
\bottomrule
\end{tabular}
\end{table}

\Cref{tab:hold,tab:noise,tab:scale} give the numbers behind the robustness figures of the main paper; nothing was retrained for \cref{tab:hold,tab:noise}.
Under a one-step observation delay \npr{} lost 24\% of its return on Walker, where Base, CAPS, ASAP and PAVE lost 11\% to 21\% and Grad-CAPS lost 43\%. The hold and delay tests measured the return of the executed policy, and \cref{cor:regret} concerns the critic value.
The observation noise followed the protocol of PAVE~\cite{lee2026stabilizing} with fixed reset seeds and noise draws, so every method saw identical perturbations.
Under reward scaling the smoothness score of \npr{} changed little, while its Lunar return fell at $c=0.1$: \cref{prop:scale} concerns the loss, and the return is not invariant.
\begin{table}[htbp]
\centering
\setlength{\tabcolsep}{4pt}
\caption{Holding actions and delaying observations at test time (TD3, 5 seeds, test reset seeds, no retraining). Return change in \% relative to the unperturbed policy with each action held for 2 or 4 control steps (h2, h4) and with a one-step observation delay (d1). Reacher and Ant were not evaluated.}
\label{tab:hold}
\footnotesize
\begin{tabular}{lcccccccccccc}
\toprule
& \multicolumn{3}{c}{Lunar} & \multicolumn{3}{c}{Pendulum} & \multicolumn{3}{c}{Hopper} & \multicolumn{3}{c}{Walker}\\
\cmidrule(lr){2-4} \cmidrule(lr){5-7} \cmidrule(lr){8-10} \cmidrule(lr){11-13}
Method & h2 & h4 & d1 & h2 & h4 & d1 & h2 & h4 & d1 & h2 & h4 & d1\\
\midrule
Base & +1 & +1 & +6 & $-$1 & $-$184 & $-$2 & $-$20 & $-$97 & $-$42 & $-$15 & $-$98 & $-$19\\
CAPS & $-$2 & $-$2 & $-$2 & $-$1 & $-$5 & $-$1 & $-$27 & $-$89 & $-$43 & $-$19 & $-$100 & $-$11\\
Grad-CAPS & +3 & +13 & +6 & $-$1 & $-$199 & $-$2 & $-$19 & $-$97 & $-$57 & $-$22 & $-$96 & $-$43\\
ASAP & +2 & $-$10 & +1 & $-$2 & $-$210 & $-$3 & $-$36 & $-$98 & $-$37 & $-$39 & $-$97 & $-$21\\
PAVE & $-$1 & 0 & $-$2 & 0 & $-$2 & $-$1 & $-$20 & $-$81 & $-$31 & $-$10 & $-$97 & $-$16\\
\npr{}, weight of \cref{tab:main} & +4 & +10 & +4 & $-$1 & $-$5 & $-$1 & +3 & $-$42 & +2 & +1 & $-$34 & $-$24\\
\npr{}, shared $\tilde\lambda=0.1$ & +1 & +6 & +5 & $-$1 & $-$5 & $-$1 & +6 & $-$27 & $-$12 & +1 & $-$34 & $-$24\\
\bottomrule
\end{tabular}
\end{table}

\begin{table}[htbp]
\centering
\setlength{\tabcolsep}{4pt}
\caption{Observation noise, protocol of PAVE~\cite{lee2026stabilizing}: $\delta\sim\mathcal U(-\sigma,\sigma)\odot\sigma_{\mathrm{base}}$ with $\sigma_{\mathrm{base}}$ the per-feature standard deviation of the observations visited by a Base policy, 20 episodes per policy with fixed reset seeds and noise draws (TD3, 5 seeds, no retraining); the protocol was run on Lunar, Hopper and Walker. }
\label{tab:noise}
\footnotesize
\begin{tabular}{lccccccccc}
\toprule
& \multicolumn{3}{c}{Lunar} & \multicolumn{3}{c}{Hopper} & \multicolumn{3}{c}{Walker}\\
\cmidrule(lr){2-4} \cmidrule(lr){5-7} \cmidrule(lr){8-10}
$\sigma$ & 0.01 & 0.05 & 0.1 & 0.01 & 0.05 & 0.1 & 0.01 & 0.05 & 0.1\\
\midrule
\multicolumn{10}{l}{\textit{Return re}}\\
Base & 208.1 & 197.7 & 170.0 & 3596 & 3594 & 3515 & 4594 & 4483 & 4530\\
CAPS & 244.1 & 232.3 & 207.8 & 2731 & 2638 & 2565 & 5183 & 5212 & 5248\\
Grad-CAPS & 229.1 & 206.3 & 196.7 & 3255 & 3168 & 2940 & 5040 & 4975 & 4807\\
ASAP & 220.2 & 215.9 & 187.3 & 2677 & 2596 & 2349 & 4817 & 4846 & 4807\\
PAVE & 257.4 & 229.1 & 184.4 & 2847 & 2613 & 2412 & 5507 & 5497 & 5496\\
\npr{}, weight of \cref{tab:main} & 225.8 & 225.5 & 219.1 & 3206 & 3207 & 3121 & 4489 & 4539 & 4519\\
\npr{}, shared $\tilde\lambda=0.1$ & 239.7 & 234.5 & 229.1 & 2681 & 2680 & 2670 & 4489 & 4539 & 4519\\
\midrule
\multicolumn{10}{l}{\textit{Smoothness score sm}}\\
Base & 1.888 & 2.285 & 2.721 & 2.722 & 3.145 & 3.562 & 1.836 & 1.978 & 2.203\\
CAPS & 0.787 & 1.142 & 1.460 & 1.713 & 1.992 & 2.321 & 1.862 & 1.945 & 2.114\\
Grad-CAPS & 0.921 & 1.378 & 1.974 & 1.387 & 1.948 & 2.396 & 1.362 & 1.534 & 1.781\\
ASAP & 1.674 & 2.130 & 2.598 & 1.895 & 2.320 & 2.605 & 1.558 & 1.712 & 1.901\\
PAVE & 0.638 & 1.150 & 2.163 & 1.100 & 1.318 & 1.564 & 1.491 & 1.640 & 1.893\\
\npr{}, weight of \cref{tab:main} & 0.120 & 0.231 & 0.454 & 0.580 & 0.773 & 1.061 & 0.367 & 0.405 & 0.484\\
\npr{}, shared $\tilde\lambda=0.1$ & 0.215 & 0.395 & 0.792 & 0.203 & 0.267 & 0.379 & 0.367 & 0.405 & 0.484\\
\bottomrule
\end{tabular}
\end{table}

\begin{table}[t]
\centering
\setlength{\tabcolsep}{3pt}
\caption{Reward scale (\cref{prop:scale}): rewards $\times c$, hyperparameters fixed (TD3, test, 5 seeds); re in unscaled units. \npr{} at the shared weight $\tilde\lambda=0.1$ (Lunar: $\lambda=0.05$, Hopper: $\lambda=0.033$, not the $0.5$ and $0.002$ of \cref{tab:main}).}
\label{tab:scale}
\scriptsize
\begin{tabular}{@{}llcccccc@{}}
\toprule
& & \multicolumn{2}{c}{$c{=}0.1$} & \multicolumn{2}{c}{$c{=}1$} & \multicolumn{2}{c}{$c{=}10$}\\
\cmidrule(lr){3-4}\cmidrule(lr){5-6}\cmidrule(l){7-8}
Task & Method & re & sm & re & sm & re & sm\\
\midrule
Lunar & PAVE & 254.8 & 0.282 & 252.0 & 0.703 & 243.5 & 1.372\\
Lunar & \npr{} & 183.0 & 0.269 & 226.0 & 0.205 & 240.1 & 0.183\\
Hopper & PAVE & 2849 & 0.532 & 2840 & 1.042 & 3222 & 1.288\\
Hopper & \npr{} & 2509 & 0.197 & 2672 & 0.204 & 3312 & 0.230\\
\bottomrule
\end{tabular}
\end{table}

\section{Shared-Weight Results}
\label{supp:shared}
\Cref{tab:shared} repeats \cref{tab:main} with \npr{} at the shared weight $\tilde\lambda=0.1$ on every task, without per-task selection on the test runs. Its mean sm factor against Base was $6.2\times$ with TD3 and $2.4\times$ with SAC, with geometric means of $4.9\times$ and $2.2\times$, and at the three-decimal precision of \cref{tab:shared} \npr{} had the lowest sm among the regularized methods on all six TD3 tasks and four SAC tasks. Its TD3 return fell below that at the per-task weight on Lunar, Reacher, Hopper and Ant, and the paired return interval against Base lay entirely below zero on TD3 Reacher and Hopper and on SAC Pendulum and Walker. The TD3 return at the shared weight was 26\% below Base on Hopper and 2\% above Base on Ant. The shared rows of \cref{tab:assump} give the critic geometry at this weight.
\begin{table}[htbp]
\centering
\setlength{\tabcolsep}{2.6pt}
\renewcommand{\arraystretch}{0.95}
\caption{\npr{} at the shared weight: \cref{tab:main} with \npr{} at $\tilde\lambda=0.1$ ($\lambda=\tilde\lambda/d$), set on validation runs (\cref{supp:results}), on every task instead of the per-task weight; all other rows as in \cref{tab:main} (test, mean over 5 training seeds, standard deviation in parentheses). Base is the unregularized reference; \textbf{bold} marks the best and \underline{underline} the second best per task and metric over all rows. \textcolor{red}{Red}: return below Base, the paired per-seed 95\% bootstrap interval of the difference lying entirely below zero. }
\label{tab:shared}
\scriptsize
\begin{tabular}{@{}lcccccccccccc@{}}
\toprule
& \multicolumn{2}{c}{Lunar} & \multicolumn{2}{c}{Pendulum} & \multicolumn{2}{c}{Reacher} & \multicolumn{2}{c}{Hopper} & \multicolumn{2}{c}{Walker} & \multicolumn{2}{c}{Ant}\\
\cmidrule(lr){2-3} \cmidrule(lr){4-5} \cmidrule(lr){6-7} \cmidrule(lr){8-9} \cmidrule(lr){10-11} \cmidrule(lr){12-13}
Method & re$\uparrow$ & sm$\downarrow$ & re$\uparrow$ & sm$\downarrow$ & re$\uparrow$ & sm$\downarrow$ & re$\uparrow$ & sm$\downarrow$ & re$\uparrow$ & sm$\downarrow$ & re$\uparrow$ & sm$\downarrow$\\
\midrule
TD3 Base & 214.0{\tiny\,(65.3)} & 1.580{\tiny\,(0.571)} & -154.5{\tiny\,(0.78)} & 1.900{\tiny\,(0.381)} & \underline{-3.36}{\tiny\,(0.08)} & 0.051{\tiny\,(0.001)} & \textbf{3595}{\tiny\,(58.3)} & 2.645{\tiny\,(0.363)} & 4596{\tiny\,(562.3)} & 1.818{\tiny\,(0.307)} & 4856{\tiny\,(1008)} & 2.158{\tiny\,(0.174)}\\
\multicolumn{13}{l}{\cellcolor{gray!12}\textit{Actor-side regularization}}\\
CAPS~\cite{mysore2021regularizing} & \underline{236.2}{\tiny\,(64.0)} & 0.715{\tiny\,(0.292)} & \textcolor{red}{-160.5}{\tiny\,(5.35)} & \underline{0.407}{\tiny\,(0.029)} & -3.37{\tiny\,(0.07)} & \underline{0.045}{\tiny\,(0.002)} & \textcolor{red}{2822}{\tiny\,(1566)} & 1.706{\tiny\,(0.628)} & \underline{5259}{\tiny\,(679.7)} & 1.848{\tiny\,(0.280)} & 4912{\tiny\,(462.4)} & 1.922{\tiny\,(0.068)}\\
Grad-CAPS~\cite{lee2024gradient} & 205.9{\tiny\,(53.0)} & 0.855{\tiny\,(0.147)} & -155.1{\tiny\,(2.28)} & 1.023{\tiny\,(0.519)} & \underline{-3.36}{\tiny\,(0.04)} & \textbf{0.040}{\tiny\,(0.001)} & \underline{3164}{\tiny\,(677.0)} & 1.273{\tiny\,(0.387)} & 5049{\tiny\,(694.8)} & \underline{1.344}{\tiny\,(0.314)} & \underline{5500}{\tiny\,(403.5)} & 1.793{\tiny\,(0.152)}\\
ASAP~\cite{kwak2026enhancingcontrolpolicysmoothness} & 207.5{\tiny\,(46.3)} & 1.694{\tiny\,(0.676)} & -154.9{\tiny\,(2.95)} & 2.283{\tiny\,(0.906)} & \textbf{-3.32}{\tiny\,(0.02)} & 0.049{\tiny\,(0.003)} & \textcolor{red}{2685}{\tiny\,(849.5)} & 1.809{\tiny\,(0.394)} & 4834{\tiny\,(540.5)} & 1.544{\tiny\,(0.221)} & 4744{\tiny\,(683.2)} & 2.110{\tiny\,(0.187)}\\
\multicolumn{13}{l}{\cellcolor{gray!12}\textit{Critic-side regularization}}\\
PAVE~\cite{lee2026stabilizing} & \textbf{252.0}{\tiny\,(19.6)} & \underline{0.703}{\tiny\,(0.269)} & \textbf{-153.6}{\tiny\,(0.86)} & 0.410{\tiny\,(0.025)} & \textcolor{red}{-3.57}{\tiny\,(0.14)} & \underline{0.045}{\tiny\,(0.004)} & 2840{\tiny\,(968.8)} & \underline{1.042}{\tiny\,(0.306)} & \textbf{5501}{\tiny\,(512.6)} & 1.492{\tiny\,(0.378)} & \textbf{5583}{\tiny\,(413.9)} & \underline{1.774}{\tiny\,(0.120)}\\
\rowcolor{blue!8}\textbf{\npr{}, shared $\tilde\lambda$} & 226.0{\tiny\,(44.2)} & \textbf{0.205}{\tiny\,(0.028)} & \underline{-154.4}{\tiny\,(0.55)} & \textbf{0.272}{\tiny\,(0.017)} & \textcolor{red}{-3.54}{\tiny\,(0.07)} & \textbf{0.040}{\tiny\,(0.002)} & \textcolor{red}{2672}{\tiny\,(766.5)} & \textbf{0.204}{\tiny\,(0.043)} & 4503{\tiny\,(621.8)} & \textbf{0.358}{\tiny\,(0.082)} & 4942{\tiny\,(892.2)} & \textbf{0.705}{\tiny\,(0.075)}\\
\midrule
SAC Base & \underline{264.6}{\tiny\,(20.8)} & 0.318{\tiny\,(0.066)} & \underline{-149.3}{\tiny\,(1.18)} & 0.510{\tiny\,(0.052)} & \textbf{-3.46}{\tiny\,(0.03)} & 0.051{\tiny\,(0.002)} & \underline{3168}{\tiny\,(435.6)} & 0.778{\tiny\,(0.053)} & \textbf{4843}{\tiny\,(513.6)} & 0.712{\tiny\,(0.100)} & \underline{5380}{\tiny\,(435.7)} & 1.994{\tiny\,(0.109)}\\
\multicolumn{13}{l}{\cellcolor{gray!12}\textit{Actor-side regularization}}\\
CAPS~\cite{mysore2021regularizing} & 216.1{\tiny\,(122.6)} & 0.284{\tiny\,(0.028)} & \textcolor{red}{-151.8}{\tiny\,(1.55)} & \underline{0.316}{\tiny\,(0.016)} & -3.51{\tiny\,(0.11)} & \underline{0.045}{\tiny\,(0.003)} & 3041{\tiny\,(966.5)} & 0.599{\tiny\,(0.135)} & 4541{\tiny\,(580.1)} & 0.667{\tiny\,(0.052)} & 5288{\tiny\,(889.8)} & 1.906{\tiny\,(0.191)}\\
Grad-CAPS~\cite{lee2024gradient} & 215.9{\tiny\,(126.6)} & 0.236{\tiny\,(0.025)} & \textbf{-148.8}{\tiny\,(0.86)} & 0.402{\tiny\,(0.017)} & \textcolor{red}{-3.53}{\tiny\,(0.08)} & \textbf{0.043}{\tiny\,(0.003)} & 2914{\tiny\,(852.0)} & 0.574{\tiny\,(0.116)} & \underline{4754}{\tiny\,(386.1)} & 0.558{\tiny\,(0.035)} & 5237{\tiny\,(922.1)} & 1.803{\tiny\,(0.101)}\\
ASAP~\cite{kwak2026enhancingcontrolpolicysmoothness} & 221.1{\tiny\,(124.4)} & \textbf{0.188}{\tiny\,(0.015)} & -149.8{\tiny\,(0.93)} & 0.379{\tiny\,(0.055)} & \textcolor{red}{-3.51}{\tiny\,(0.03)} & \textbf{0.043}{\tiny\,(0.003)} & \textbf{3297}{\tiny\,(378.0)} & \underline{0.469}{\tiny\,(0.061)} & 4392{\tiny\,(734.4)} & \underline{0.529}{\tiny\,(0.059)} & 4659{\tiny\,(2422)} & \underline{1.520}{\tiny\,(0.205)}\\
\multicolumn{13}{l}{\cellcolor{gray!12}\textit{Critic-side regularization}}\\
PAVE~\cite{lee2026stabilizing} & 264.2{\tiny\,(12.5)} & \underline{0.200}{\tiny\,(0.015)} & \textcolor{red}{-151.0}{\tiny\,(0.64)} & 0.347{\tiny\,(0.025)} & \underline{-3.48}{\tiny\,(0.17)} & 0.051{\tiny\,(0.002)} & 3062{\tiny\,(863.2)} & 0.549{\tiny\,(0.090)} & \textcolor{red}{4545}{\tiny\,(388.9)} & 0.568{\tiny\,(0.068)} & \textbf{5502}{\tiny\,(384.1)} & 1.628{\tiny\,(0.126)}\\
\rowcolor{blue!8}\textbf{\npr{}, shared $\tilde\lambda$} & \textbf{267.2}{\tiny\,(9.76)} & 0.221{\tiny\,(0.030)} & \textcolor{red}{-165.7}{\tiny\,(2.49)} & \textbf{0.246}{\tiny\,(0.017)} & \underline{-3.48}{\tiny\,(0.07)} & 0.047{\tiny\,(0.001)} & 2698{\tiny\,(647.4)} & \textbf{0.184}{\tiny\,(0.019)} & \textcolor{red}{4504}{\tiny\,(481.4)} & \textbf{0.271}{\tiny\,(0.056)} & 5036{\tiny\,(112.5)} & \textbf{0.638}{\tiny\,(0.059)}\\
\bottomrule
\end{tabular}
\end{table}

\section{Time-Domain Action Increments}
\label{supp:rmsda}
\Cref{tab:rmsda} measures smoothness in the time domain on the saved test trajectories of \cref{tab:main}, without a new rollout: the root mean square of $\|a_{t+1}-a_t\|_2$ within each episode, averaged over the ten test episodes. With TD3, \npr{} had the lowest increment on all six tasks, $9.4\times$ below Base on average and $5.4\times$ in geometric mean, against $3.8\times$ and $2.2\times$ for PAVE. With SAC it had the lowest increment on Pendulum, Hopper, Walker and Ant, the four tasks on which it had the lowest sm; ASAP and PAVE were lower on Lunar, and ASAP and Grad-CAPS on Reacher.
\begin{table}[htbp]
\centering
\setlength{\tabcolsep}{3pt}
\caption{Root mean square of the action increment $\|a_{t+1}-a_t\|_2$ on the test trajectories of \cref{tab:main} (5 seeds, 10 episodes per seed, increments within an episode, std over seeds in parentheses). Bold: lowest.}
\label{tab:rmsda}
\footnotesize
\begin{tabular}{@{}lcccccc@{}}
\toprule
Method & Lunar & Pendulum & Reacher & Hopper & Walker & Ant\\
\midrule
\multicolumn{7}{l}{\textit{TD3}}\\
Base & 0.946{\tiny\,(0.224)} & 1.86{\tiny\,(1.28)} & 0.0223{\tiny\,(0.0006)} & 0.952{\tiny\,(0.117)} & 0.907{\tiny\,(0.148)} & 1.32{\tiny\,(0.10)}\\
CAPS & 0.383{\tiny\,(0.109)} & 0.132{\tiny\,(0.014)} & 0.0215{\tiny\,(0.0014)} & 0.963{\tiny\,(0.548)} & 0.905{\tiny\,(0.127)} & 1.24{\tiny\,(0.10)}\\
Grad-CAPS & 0.452{\tiny\,(0.031)} & 0.668{\tiny\,(0.577)} & 0.0230{\tiny\,(0.0018)} & 0.497{\tiny\,(0.111)} & 0.675{\tiny\,(0.132)} & 1.13{\tiny\,(0.08)}\\
ASAP & 0.878{\tiny\,(0.239)} & 1.74{\tiny\,(1.17)} & 0.0225{\tiny\,(0.0016)} & 0.756{\tiny\,(0.130)} & 0.757{\tiny\,(0.104)} & 1.33{\tiny\,(0.09)}\\
PAVE & 0.388{\tiny\,(0.101)} & 0.126{\tiny\,(0.012)} & 0.0209{\tiny\,(0.0016)} & 0.456{\tiny\,(0.115)} & 0.745{\tiny\,(0.158)} & 1.15{\tiny\,(0.07)}\\
\npr{} & \textbf{0.0448}{\tiny\,(0.0047)} & \textbf{0.0800}{\tiny\,(0.0037)} & \textbf{0.0195}{\tiny\,(0.0015)} & \textbf{0.238}{\tiny\,(0.031)} & \textbf{0.202}{\tiny\,(0.044)} & \textbf{0.547}{\tiny\,(0.055)}\\
\midrule
\multicolumn{7}{l}{\textit{SAC}}\\
Base & 0.157{\tiny\,(0.014)} & 0.193{\tiny\,(0.021)} & 0.0231{\tiny\,(0.0013)} & 0.302{\tiny\,(0.018)} & 0.374{\tiny\,(0.057)} & 1.25{\tiny\,(0.05)}\\
CAPS & 0.129{\tiny\,(0.033)} & 0.108{\tiny\,(0.011)} & 0.0215{\tiny\,(0.0009)} & 0.251{\tiny\,(0.038)} & 0.347{\tiny\,(0.039)} & 1.22{\tiny\,(0.06)}\\
Grad-CAPS & 0.110{\tiny\,(0.030)} & 0.144{\tiny\,(0.014)} & 0.0205{\tiny\,(0.0001)} & 0.252{\tiny\,(0.044)} & 0.314{\tiny\,(0.028)} & 1.15{\tiny\,(0.06)}\\
ASAP & \textbf{0.0872}{\tiny\,(0.0273)} & 0.132{\tiny\,(0.025)} & \textbf{0.0201}{\tiny\,(0.0006)} & 0.206{\tiny\,(0.025)} & 0.303{\tiny\,(0.021)} & 0.994{\tiny\,(0.071)}\\
PAVE & 0.0956{\tiny\,(0.0071)} & 0.122{\tiny\,(0.009)} & 0.0228{\tiny\,(0.0012)} & 0.235{\tiny\,(0.029)} & 0.309{\tiny\,(0.029)} & 1.11{\tiny\,(0.03)}\\
\npr{} & 0.110{\tiny\,(0.010)} & \textbf{0.0931}{\tiny\,(0.0047)} & 0.0210{\tiny\,(0.0012)} & \textbf{0.0941}{\tiny\,(0.0177)} & \textbf{0.200}{\tiny\,(0.048)} & \textbf{0.497}{\tiny\,(0.020)}\\
\bottomrule
\end{tabular}
\end{table}

\section{Learning Curves, Training Time and Memory}
\label{supp:curves}
\begin{figure}[htbp]
\centering
\includegraphics[width=\textwidth]{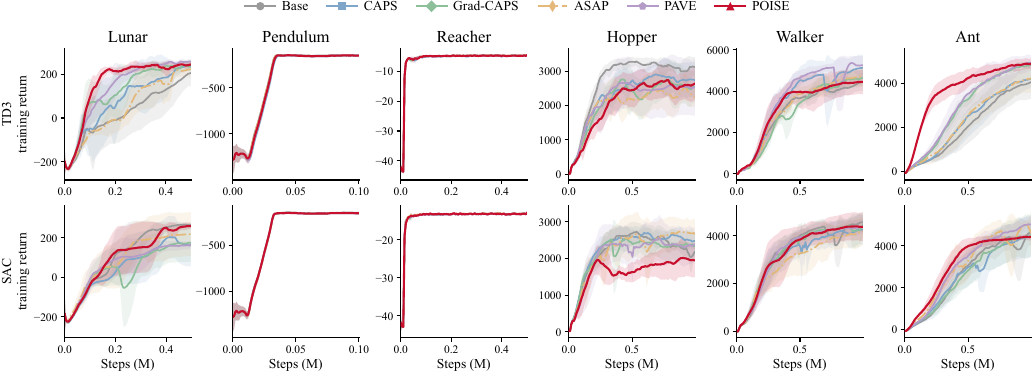}
\caption{Learning curves of the runs of \cref{tab:main} with TD3 (top) and SAC (bottom): mean training return over 5 seeds, with the standard deviation shaded.}
\label{fig:curves}
\end{figure}

\Cref{fig:curves} plots the training return of the runs of \cref{tab:main}.
\npr{} changed critic training and added no operation to policy inference.

\paragraph{Complexity.}
Let $C$ and $M$ be the time and the memory of one forward and backward pass of the critic on a batch of $N$ samples, so that one update of Base costs $\mathcal O(C)$ time and $\mathcal O(M)$ memory. The implementation evaluated three action gradients, built the Hessian with $d$ backward passes, and performed two eigendecompositions and two linear solves per sample. One update of \npr{} therefore costs $\mathcal O(dC+Nd^3)$ time and $\mathcal O(dM+Nd^2)$ memory for each regularized critic head, which is one head with TD3 and two with SAC. PAVE uses finite differences and a Hutchinson estimator~\cite{lee2026stabilizing}, so its number of critic passes is independent of $d$ and its cost is $\mathcal O(C)$ time and $\mathcal O(M)$ memory. The measurements in \cref{tab:cost} followed this difference. The update time of \npr{} rose from 11.2 ms at $d=1$ to 23.5 ms at $d=8$, and the multiple of PAVE over Base stayed between 3.6 and 3.8.

\paragraph{Measured cost.}
\Cref{tab:cost} times the update alone at batch size 256, and \cref{tab:traintime} times a full environment step of the training runs with each run timed alone, at batch size 256 with TD3 and 100 with SAC. With TD3, \npr{} took 2.8 to 4.7 times the time of Base, against 3.1 to 3.7 times for PAVE, and was faster than PAVE on Lunar, Pendulum, Reacher and Hopper. With SAC it took 2.9 to 5.9 times, against 2.2 to 2.5 times for PAVE, because the penalty is computed on both critic heads.
\Cref{tab:memory} lists the peak GPU memory of the TD3 and SAC updates. \npr{} allocated 49 to 94 MB on the six tasks, against 69 to 74 MB for PAVE and 27 to 30 MB for Base, and stayed below PAVE on Lunar, Pendulum, Reacher and Hopper. Its memory grew with the action dimension. With SAC at its training batch of 100, \npr{} allocated 33 to 59 MB on the six tasks against 33 to 35 MB for PAVE and 22 to 24 MB for Base. In the same benchmark the SAC update of \npr{} took 3.2 to 6.3 times the time of Base, against 2.6 to 2.7 times for PAVE.
\begin{table}[htbp]
\centering
\setlength{\tabcolsep}{5pt}
\caption{Training time per environment step in milliseconds, including the environment step and one update. Each run was timed alone on one RTX A5000 over 19{,}500 steps after the warm-up, with one seed.}
\label{tab:traintime}
\footnotesize
\begin{tabular}{@{}lcccccc@{}}
\toprule
Method & Lunar & Pendulum & Reacher & Hopper & Walker & Ant\\
\midrule
\multicolumn{7}{l}{\textit{TD3}}\\
Base & 6.3 & 4.7 & 5.0 & 5.1 & 5.0 & 5.5\\
CAPS & 7.9 & 5.5 & 5.8 & 7.0 & 6.0 & 6.5\\
Grad-CAPS & 7.8 & 5.7 & 5.8 & 6.8 & 6.7 & 6.5\\
ASAP & 8.7 & 8.0 & 7.1 & 7.2 & 7.9 & 7.8\\
PAVE & 19.7 & 17.3 & 16.1 & 17.9 & 16.9 & 16.9\\
\npr{} & 17.5 & 13.5 & 14.5 & 16.3 & 21.9 & 25.9\\
\midrule
\multicolumn{7}{l}{\textit{SAC}}\\
Base & 9.1 & 7.9 & 8.4 & 8.0 & 8.1 & 8.5\\
CAPS & 11.6 & 10.5 & 10.6 & 10.5 & 10.5 & 10.7\\
Grad-CAPS & 11.9 & 10.5 & 10.6 & 10.5 & 10.5 & 11.4\\
ASAP & 13.7 & 12.8 & 13.0 & 12.9 & 12.9 & 14.4\\
PAVE & 20.5 & 19.2 & 19.4 & 19.3 & 19.4 & 21.4\\
\npr{} & 28.5 & 23.2 & 27.0 & 30.6 & 41.3 & 50.5\\
\bottomrule
\end{tabular}
\end{table}

\begin{table}[htbp]
\centering
\setlength{\tabcolsep}{5pt}
\caption{Peak GPU memory in MB allocated by PyTorch during 2000 training updates on one RTX A5000: TD3 at batch size 256, the runs of \cref{tab:cost}, and SAC at its training batch size of 100.}
\label{tab:memory}
\footnotesize
\begin{tabular}{@{}lcccccc@{}}
\toprule
Method & Lunar & Pendulum & Reacher & Hopper & Walker & Ant\\
\midrule
\multicolumn{7}{l}{\textit{TD3, batch 256}}\\
Base & 27 & 27 & 27 & 27 & 27 & 30\\
CAPS & 31 & 31 & 31 & 31 & 32 & 35\\
Grad-CAPS & 31 & 31 & 31 & 31 & 32 & 35\\
ASAP & 34 & 34 & 34 & 35 & 35 & 38\\
PAVE & 70 & 69 & 70 & 70 & 70 & 74\\
\npr{} & 55 & 49 & 55 & 61 & 79 & 94\\
\midrule
\multicolumn{7}{l}{\textit{SAC, batch 100}}\\
Base & 22 & 22 & 22 & 22 & 22 & 24\\
CAPS & 23 & 23 & 23 & 23 & 24 & 25\\
Grad-CAPS & 23 & 23 & 23 & 23 & 24 & 25\\
ASAP & 25 & 25 & 25 & 25 & 25 & 27\\
PAVE & 33 & 33 & 33 & 33 & 33 & 35\\
\npr{} & 37 & 33 & 37 & 40 & 50 & 59\\
\bottomrule
\end{tabular}
\end{table}

\FloatBarrier
\section{Weight Sweeps}
\label{supp:sweeps}
\Cref{tab:frontier} raises the weights of PAVE and Grad-CAPS to 3 and 10 times their published values on the training seeds of \cref{tab:main}. The sm of Grad-CAPS fell with its weight on every task. At 10 times its weight it reached 0.027 on Reacher against 0.044 for \npr{}, and 0.576 on Hopper with a return of 3074 against 0.561 and 3208 for \npr{}. On the other four tasks the lowest sm of either baseline stayed above that of \npr{} by 3.7 times on Lunar, 1.4 times on Pendulum, 2.5 times on Walker and 1.5 times on Ant. Higher weights of PAVE lowered its sm by at most 1.7 times, and its sm at some raised weight exceeded that at the published weights on Walker and Ant.
\Cref{tab:sqshift} doubles the weight of the temporal-only form of \cref{tab:ablation}. The doubled form had a higher sm than \npr{} on all three tasks, with 0.213, 0.233 and 0.437 against 0.205, 0.204 and 0.358.
\begin{table*}[htbp]
\centering
\setlength{\tabcolsep}{1.9pt}
\caption{Baseline weights raised to 3 and 10 times their published values (TD3, test, the 5 training seeds of \cref{tab:main}, standard deviation in parentheses). PAVE scales all three weights, Grad-CAPS its weight $\lambda_T$. The published rows and \npr{} are those of \cref{tab:main}.}
\label{tab:frontier}
\scriptsize
\begin{tabular}{@{}lcccccccccccc@{}}
\toprule
& \multicolumn{2}{c}{Lunar} & \multicolumn{2}{c}{Pendulum} & \multicolumn{2}{c}{Reacher} & \multicolumn{2}{c}{Hopper} & \multicolumn{2}{c}{Walker} & \multicolumn{2}{c}{Ant}\\
\cmidrule(lr){2-3}\cmidrule(lr){4-5}\cmidrule(lr){6-7}\cmidrule(lr){8-9}\cmidrule(lr){10-11}\cmidrule(lr){12-13}
Method & re$\uparrow$ & sm$\downarrow$ & re$\uparrow$ & sm$\downarrow$ & re$\uparrow$ & sm$\downarrow$ & re$\uparrow$ & sm$\downarrow$ & re$\uparrow$ & sm$\downarrow$ & re$\uparrow$ & sm$\downarrow$\\
\midrule
Base & 214.0{\tiny\,(65.3)} & 1.580{\tiny\,(0.571)} & -154.5{\tiny\,(0.78)} & 1.900{\tiny\,(0.381)} & -3.36{\tiny\,(0.08)} & 0.051{\tiny\,(0.001)} & 3595{\tiny\,(58.3)} & 2.645{\tiny\,(0.363)} & 4596{\tiny\,(562.3)} & 1.818{\tiny\,(0.307)} & 4856{\tiny\,(1008)} & 2.158{\tiny\,(0.174)}\\
\midrule
Grad-CAPS, published weights & 205.9{\tiny\,(53.0)} & 0.855{\tiny\,(0.147)} & -155.1{\tiny\,(2.28)} & 1.023{\tiny\,(0.519)} & -3.36{\tiny\,(0.04)} & 0.040{\tiny\,(0.001)} & 3164{\tiny\,(677.0)} & 1.273{\tiny\,(0.387)} & 5049{\tiny\,(694.8)} & 1.344{\tiny\,(0.314)} & 5500{\tiny\,(403.5)} & 1.793{\tiny\,(0.152)}\\
Grad-CAPS, weights $\times3$ & 256.8{\tiny\,(15.7)} & 0.519{\tiny\,(0.061)} & -154.2{\tiny\,(2.96)} & 0.569{\tiny\,(0.160)} & -3.36{\tiny\,(0.03)} & 0.034{\tiny\,(0.001)} & 3402{\tiny\,(272.4)} & 0.840{\tiny\,(0.092)} & 4895{\tiny\,(628.8)} & 1.341{\tiny\,(0.180)} & 4987{\tiny\,(908.3)} & 1.540{\tiny\,(0.071)}\\
Grad-CAPS, weights $\times10$ & 242.2{\tiny\,(27.0)} & 0.457{\tiny\,(0.047)} & -153.2{\tiny\,(1.97)} & 0.422{\tiny\,(0.015)} & -3.44{\tiny\,(0.06)} & 0.027{\tiny\,(0.002)} & 3074{\tiny\,(925.4)} & 0.576{\tiny\,(0.082)} & 5069{\tiny\,(581.6)} & 0.909{\tiny\,(0.200)} & 5342{\tiny\,(800.7)} & 1.095{\tiny\,(0.040)}\\
\midrule
PAVE, published weights & 252.0{\tiny\,(19.6)} & 0.703{\tiny\,(0.269)} & -153.6{\tiny\,(0.86)} & 0.410{\tiny\,(0.025)} & -3.57{\tiny\,(0.14)} & 0.045{\tiny\,(0.004)} & 2840{\tiny\,(968.8)} & 1.042{\tiny\,(0.306)} & 5501{\tiny\,(512.6)} & 1.492{\tiny\,(0.378)} & 5583{\tiny\,(413.9)} & 1.774{\tiny\,(0.120)}\\
PAVE, weights $\times3$ & 251.1{\tiny\,(6.50)} & 0.498{\tiny\,(0.036)} & -157.5{\tiny\,(3.41)} & 0.403{\tiny\,(0.051)} & -3.61{\tiny\,(0.11)} & 0.044{\tiny\,(0.005)} & 3057{\tiny\,(652.3)} & 0.928{\tiny\,(0.145)} & 5139{\tiny\,(443.2)} & 1.577{\tiny\,(0.437)} & 5021{\tiny\,(1161)} & 1.752{\tiny\,(0.091)}\\
PAVE, weights $\times10$ & 258.3{\tiny\,(8.11)} & 0.412{\tiny\,(0.111)} & -157.1{\tiny\,(2.36)} & 0.372{\tiny\,(0.029)} & -3.98{\tiny\,(0.13)} & 0.037{\tiny\,(0.003)} & 2877{\tiny\,(580.7)} & 0.946{\tiny\,(0.134)} & 5225{\tiny\,(1014)} & 1.385{\tiny\,(0.399)} & 4901{\tiny\,(1177)} & 1.968{\tiny\,(0.117)}\\
\npr{}, weight of \cref{tab:main} & 239.8{\tiny\,(33.9)} & 0.110{\tiny\,(0.011)} & -154.4{\tiny\,(0.55)} & 0.272{\tiny\,(0.017)} & -3.36{\tiny\,(0.04)} & 0.044{\tiny\,(0.001)} & 3208{\tiny\,(623.1)} & 0.561{\tiny\,(0.101)} & 4503{\tiny\,(621.8)} & 0.358{\tiny\,(0.082)} & 5559{\tiny\,(292.4)} & 0.726{\tiny\,(0.092)}\\
\bottomrule
\end{tabular}
\end{table*}

\begin{table}[htbp]
\centering
\setlength{\tabcolsep}{2.2pt}
\caption{Temporal-only form of \cref{tab:ablation} at the nominal and the doubled weight (TD3, test, 5 seeds, standard deviation in parentheses). The nominal weight is the shared weight of \npr{}: $\lambda=0.05$ on Lunar, $\tilde\lambda=0.1$ on Hopper and Walker. The row $\times1$ is that of \cref{tab:ablation}.}
\label{tab:sqshift}
\scriptsize
\begin{tabular}{@{}lcccccc@{}}
\toprule
& \multicolumn{2}{c}{Lunar} & \multicolumn{2}{c}{Hopper} & \multicolumn{2}{c}{Walker}\\
\cmidrule(lr){2-3}\cmidrule(lr){4-5}\cmidrule(lr){6-7}
Variant & re$\uparrow$ & sm$\downarrow$ & re$\uparrow$ & sm$\downarrow$ & re$\uparrow$ & sm$\downarrow$\\
\midrule
\npr{}, shared weight & 226.0{\tiny\,(44.2)} & 0.205{\tiny\,(0.028)} & 2672{\tiny\,(766.5)} & 0.204{\tiny\,(0.043)} & 4503{\tiny\,(621.8)} & 0.358{\tiny\,(0.082)}\\
\midrule
Temporal only, weight $\times1$ & 219.6{\tiny\,(53.7)} & 0.279{\tiny\,(0.081)} & 2305{\tiny\,(1047)} & 0.333{\tiny\,(0.068)} & 4517{\tiny\,(391.2)} & 0.517{\tiny\,(0.128)}\\
Temporal only, weight $\times2$ & 240.0{\tiny\,(32.6)} & 0.213{\tiny\,(0.038)} & 2380{\tiny\,(829.3)} & 0.233{\tiny\,(0.051)} & 4312{\tiny\,(879.8)} & 0.437{\tiny\,(0.061)}\\
\bottomrule
\end{tabular}
\end{table}

\section{Additional Training Seeds}
\label{supp:seeds}
\Cref{tab:newseeds} repeated the TD3 comparisons on five further training seeds unused for weight selection. Both \npr{} settings had lower mean sm than every baseline on Lunar, Pendulum, Hopper, Walker and Ant. On Reacher, the shared setting tied with Grad-CAPS at the displayed precision of 0.040. On these five seeds alone, a two-sided paired $t$-test over seeds gave $p\le0.030$ for the sm of \npr{} at the per-task weight against each of the five other methods on these five tasks. At the per-task weight, Hopper return was 3504 against 2692 for Base, and Walker return was 3718 against 4092.
\begin{table*}[htbp]
\centering
\setlength{\tabcolsep}{1.9pt}
\caption{TD3 on five further training seeds that no weight selection used (test, standard deviation in parentheses). Per-task weights were from \cref{tab:main}; shared weights used $\lambda=0.1/d$, coinciding on Pendulum and Walker.}
\label{tab:newseeds}
\scriptsize
\begin{tabular}{@{}lcccccccccccc@{}}
\toprule
& \multicolumn{2}{c}{Lunar} & \multicolumn{2}{c}{Pendulum} & \multicolumn{2}{c}{Reacher} & \multicolumn{2}{c}{Hopper} & \multicolumn{2}{c}{Walker} & \multicolumn{2}{c}{Ant}\\
\cmidrule(lr){2-3}\cmidrule(lr){4-5}\cmidrule(lr){6-7}\cmidrule(lr){8-9}\cmidrule(lr){10-11}\cmidrule(lr){12-13}
Method & re$\uparrow$ & sm$\downarrow$ & re$\uparrow$ & sm$\downarrow$ & re$\uparrow$ & sm$\downarrow$ & re$\uparrow$ & sm$\downarrow$ & re$\uparrow$ & sm$\downarrow$ & re$\uparrow$ & sm$\downarrow$\\
\midrule
Base & 173.3{\tiny\,(122.4)} & 2.171{\tiny\,(0.951)} & -154.9{\tiny\,(2.53)} & 1.324{\tiny\,(0.085)} & -3.35{\tiny\,(0.05)} & 0.051{\tiny\,(0.001)} & 2692{\tiny\,(842.8)} & 2.248{\tiny\,(0.333)} & 4092{\tiny\,(1131)} & 1.928{\tiny\,(0.506)} & 5084{\tiny\,(486.2)} & 2.084{\tiny\,(0.091)}\\
\midrule
CAPS & 242.6{\tiny\,(20.5)} & 0.752{\tiny\,(0.115)} & -161.7{\tiny\,(2.94)} & 0.434{\tiny\,(0.037)} & -3.33{\tiny\,(0.04)} & 0.048{\tiny\,(0.003)} & 3548{\tiny\,(109.3)} & 1.807{\tiny\,(0.321)} & 5353{\tiny\,(546.4)} & 1.693{\tiny\,(0.233)} & 4339{\tiny\,(848.6)} & 2.248{\tiny\,(0.159)}\\
Grad-CAPS & 238.4{\tiny\,(16.9)} & 0.810{\tiny\,(0.048)} & -154.9{\tiny\,(3.91)} & 0.732{\tiny\,(0.240)} & -3.39{\tiny\,(0.06)} & 0.040{\tiny\,(0.002)} & 2985{\tiny\,(692.0)} & 1.351{\tiny\,(0.142)} & 4896{\tiny\,(356.8)} & 1.348{\tiny\,(0.250)} & 4979{\tiny\,(445.3)} & 1.823{\tiny\,(0.053)}\\
ASAP & 255.4{\tiny\,(10.6)} & 1.548{\tiny\,(0.687)} & -153.5{\tiny\,(0.83)} & 1.355{\tiny\,(0.390)} & -3.31{\tiny\,(0.06)} & 0.049{\tiny\,(0.001)} & 3327{\tiny\,(646.2)} & 1.991{\tiny\,(0.308)} & 4436{\tiny\,(673.5)} & 1.464{\tiny\,(0.194)} & 5352{\tiny\,(497.9)} & 2.014{\tiny\,(0.194)}\\
PAVE & 215.4{\tiny\,(60.0)} & 0.905{\tiny\,(0.541)} & -153.5{\tiny\,(1.05)} & 0.430{\tiny\,(0.022)} & -3.46{\tiny\,(0.06)} & 0.045{\tiny\,(0.001)} & 3148{\tiny\,(456.2)} & 1.103{\tiny\,(0.202)} & 5078{\tiny\,(601.4)} & 1.657{\tiny\,(0.280)} & 5037{\tiny\,(1448)} & 1.825{\tiny\,(0.220)}\\
\midrule
\npr{}, weight of \cref{tab:main} & 222.4{\tiny\,(43.3)} & 0.111{\tiny\,(0.011)} & -157.3{\tiny\,(4.67)} & 0.288{\tiny\,(0.029)} & -3.39{\tiny\,(0.05)} & 0.044{\tiny\,(0.002)} & 3504{\tiny\,(76.8)} & 0.641{\tiny\,(0.121)} & 3718{\tiny\,(329.4)} & 0.311{\tiny\,(0.027)} & 5099{\tiny\,(936.6)} & 0.680{\tiny\,(0.076)}\\
\npr{}, shared weight & 208.7{\tiny\,(52.1)} & 0.208{\tiny\,(0.016)} & -157.3{\tiny\,(4.67)} & 0.288{\tiny\,(0.029)} & -3.49{\tiny\,(0.04)} & 0.040{\tiny\,(0.001)} & 3357{\tiny\,(149.1)} & 0.194{\tiny\,(0.053)} & 3718{\tiny\,(329.4)} & 0.311{\tiny\,(0.027)} & 5223{\tiny\,(408.1)} & 0.640{\tiny\,(0.058)}\\
\bottomrule
\end{tabular}
\end{table*}

\Cref{tab:newseedssac} repeated the SAC comparisons on the same five further seeds. \npr{} at the per-task weight had lower mean sm than every baseline on Pendulum, Hopper, Walker and Ant. These are the four tasks on which \npr{} had the lowest sm in \cref{tab:main}. On these five seeds alone, the paired $t$-test gave $p\le0.009$ for the sm of \npr{} against each of the five other methods on Hopper, Walker and Ant. On Lunar and Reacher, the lowest scores were 0.190 for ASAP and 0.039 for Grad-CAPS against 0.214 and 0.044 for \npr{}. Hopper return was 2876 against 3130 for Base, and Ant return was 5017 against 5204.
\begin{table*}[htbp]
\centering
\setlength{\tabcolsep}{1.9pt}
\caption{SAC on the five further training seeds of \cref{tab:newseeds} (test, standard deviation in parentheses). Per-task weights were from \cref{tab:main}; shared weights used $\lambda=0.1/d$, differing on Pendulum and Walker only.}
\label{tab:newseedssac}
\scriptsize
\begin{tabular}{@{}lcccccccccccc@{}}
\toprule
& \multicolumn{2}{c}{Lunar} & \multicolumn{2}{c}{Pendulum} & \multicolumn{2}{c}{Reacher} & \multicolumn{2}{c}{Hopper} & \multicolumn{2}{c}{Walker} & \multicolumn{2}{c}{Ant}\\
\cmidrule(lr){2-3}\cmidrule(lr){4-5}\cmidrule(lr){6-7}\cmidrule(lr){8-9}\cmidrule(lr){10-11}\cmidrule(lr){12-13}
Method & re$\uparrow$ & sm$\downarrow$ & re$\uparrow$ & sm$\downarrow$ & re$\uparrow$ & sm$\downarrow$ & re$\uparrow$ & sm$\downarrow$ & re$\uparrow$ & sm$\downarrow$ & re$\uparrow$ & sm$\downarrow$\\
\midrule
Base & 205.4{\tiny\,(124.8)} & 0.421{\tiny\,(0.181)} & -149.3{\tiny\,(1.24)} & 0.539{\tiny\,(0.051)} & -3.48{\tiny\,(0.04)} & 0.049{\tiny\,(0.002)} & 3130{\tiny\,(367.7)} & 0.734{\tiny\,(0.099)} & 4722{\tiny\,(264.4)} & 0.845{\tiny\,(0.217)} & 5204{\tiny\,(240.1)} & 1.868{\tiny\,(0.171)}\\
\midrule
CAPS & 265.9{\tiny\,(15.7)} & 0.308{\tiny\,(0.099)} & -152.9{\tiny\,(2.43)} & 0.331{\tiny\,(0.046)} & -3.55{\tiny\,(0.09)} & 0.044{\tiny\,(0.002)} & 3345{\tiny\,(227.3)} & 0.641{\tiny\,(0.079)} & 4713{\tiny\,(511.8)} & 0.736{\tiny\,(0.206)} & 5549{\tiny\,(420.7)} & 1.877{\tiny\,(0.234)}\\
Grad-CAPS & 215.9{\tiny\,(119.9)} & 0.251{\tiny\,(0.043)} & -148.5{\tiny\,(0.87)} & 0.393{\tiny\,(0.032)} & -3.57{\tiny\,(0.09)} & 0.039{\tiny\,(0.001)} & 3075{\tiny\,(891.6)} & 0.495{\tiny\,(0.062)} & 4902{\tiny\,(339.1)} & 0.614{\tiny\,(0.077)} & 5547{\tiny\,(342.7)} & 1.653{\tiny\,(0.118)}\\
ASAP & 218.9{\tiny\,(125.2)} & 0.190{\tiny\,(0.018)} & -148.4{\tiny\,(0.88)} & 0.358{\tiny\,(0.020)} & -3.52{\tiny\,(0.09)} & 0.042{\tiny\,(0.002)} & 2507{\tiny\,(917.2)} & 0.426{\tiny\,(0.053)} & 4898{\tiny\,(258.6)} & 0.567{\tiny\,(0.047)} & 4789{\tiny\,(2727)} & 1.423{\tiny\,(0.174)}\\
PAVE & 264.7{\tiny\,(4.70)} & 0.206{\tiny\,(0.024)} & -150.8{\tiny\,(0.40)} & 0.334{\tiny\,(0.018)} & -3.46{\tiny\,(0.04)} & 0.049{\tiny\,(0.003)} & 3155{\tiny\,(754.4)} & 0.610{\tiny\,(0.083)} & 4728{\tiny\,(383.6)} & 0.561{\tiny\,(0.056)} & 5212{\tiny\,(272.3)} & 1.575{\tiny\,(0.194)}\\
\midrule
\npr{}, weight of \cref{tab:main} & 269.9{\tiny\,(8.99)} & 0.214{\tiny\,(0.033)} & -151.6{\tiny\,(0.24)} & 0.275{\tiny\,(0.008)} & -3.53{\tiny\,(0.09)} & 0.044{\tiny\,(0.002)} & 2876{\tiny\,(760.7)} & 0.191{\tiny\,(0.038)} & 4777{\tiny\,(250.7)} & 0.313{\tiny\,(0.065)} & 5017{\tiny\,(269.8)} & 0.619{\tiny\,(0.062)}\\
\npr{}, shared weight & 269.9{\tiny\,(8.99)} & 0.214{\tiny\,(0.033)} & -165.2{\tiny\,(1.63)} & 0.226{\tiny\,(0.012)} & -3.53{\tiny\,(0.09)} & 0.044{\tiny\,(0.002)} & 2876{\tiny\,(760.7)} & 0.191{\tiny\,(0.038)} & 4659{\tiny\,(383.0)} & 0.273{\tiny\,(0.058)} & 5017{\tiny\,(269.8)} & 0.619{\tiny\,(0.062)}\\
\bottomrule
\end{tabular}
\end{table*}

\section{Automatic Entropy Tuning}
\label{supp:sacauto}
\Cref{tab:sacauto} repeated the SAC runs of Base, PAVE and \npr{} from \cref{tab:main} with automatic entropy tuning on the same five seeds. Every method kept its weight from \cref{tab:main}, which had been set at the fixed entropy coefficient. \npr{} had the lowest sm of the three methods on all six tasks. A two-sided paired $t$-test over seeds gave $p\le0.011$ for the sm of \npr{} against Base and against PAVE on every task. The return of \npr{} was 2697 against 3069 for Base on Hopper and 4403 against 4939 on Ant.
\begin{table*}[htbp]
\centering
\setlength{\tabcolsep}{2.6pt}
\caption{SAC with automatic entropy tuning for Base, PAVE and \npr{} (test, the five training seeds of \cref{tab:main}, standard deviation in parentheses). The weights of every method were those of \cref{tab:main}, which were set at the fixed entropy coefficient.}
\label{tab:sacauto}
\scriptsize
\begin{tabular}{@{}lcccccccccccc@{}}
\toprule
& \multicolumn{2}{c}{Lunar} & \multicolumn{2}{c}{Pendulum} & \multicolumn{2}{c}{Reacher} & \multicolumn{2}{c}{Hopper} & \multicolumn{2}{c}{Walker} & \multicolumn{2}{c}{Ant}\\
\cmidrule(lr){2-3}\cmidrule(lr){4-5}\cmidrule(lr){6-7}\cmidrule(lr){8-9}\cmidrule(lr){10-11}\cmidrule(lr){12-13}
Method & re$\uparrow$ & sm$\downarrow$ & re$\uparrow$ & sm$\downarrow$ & re$\uparrow$ & sm$\downarrow$ & re$\uparrow$ & sm$\downarrow$ & re$\uparrow$ & sm$\downarrow$ & re$\uparrow$ & sm$\downarrow$\\
\midrule
Base & 279.2{\tiny\,(4.90)} & 0.435{\tiny\,(0.039)} & -148.1{\tiny\,(1.25)} & 0.744{\tiny\,(0.145)} & -3.35{\tiny\,(0.04)} & 0.052{\tiny\,(0.002)} & 3069{\tiny\,(620.1)} & 1.137{\tiny\,(0.161)} & 4461{\tiny\,(431.2)} & 0.989{\tiny\,(0.163)} & 4939{\tiny\,(867.8)} & 2.051{\tiny\,(0.181)}\\
PAVE & 282.8{\tiny\,(2.60)} & 0.241{\tiny\,(0.021)} & -151.7{\tiny\,(3.12)} & 0.370{\tiny\,(0.040)} & -3.44{\tiny\,(0.04)} & 0.052{\tiny\,(0.003)} & 1877{\tiny\,(827.9)} & 0.555{\tiny\,(0.067)} & 4720{\tiny\,(652.7)} & 0.695{\tiny\,(0.084)} & 5217{\tiny\,(665.0)} & 1.777{\tiny\,(0.154)}\\
\npr{} & 280.1{\tiny\,(3.11)} & 0.161{\tiny\,(0.019)} & -153.5{\tiny\,(4.09)} & 0.304{\tiny\,(0.056)} & -3.57{\tiny\,(0.06)} & 0.038{\tiny\,(0.002)} & 2697{\tiny\,(846.7)} & 0.141{\tiny\,(0.012)} & 4612{\tiny\,(704.2)} & 0.318{\tiny\,(0.046)} & 4403{\tiny\,(222.8)} & 0.488{\tiny\,(0.064)}\\
\bottomrule
\end{tabular}
\end{table*}

\section{Auxiliary Properties}
\label{supp:proofs}
\subsection{The saddle-free surrogate}
\label{supp:surrogate}
For symmetric $B=V\mathrm{diag}(\lambda)V^\top$, the metric $|B|=V\mathrm{diag}(\max(|\lambda_i|,\eta\kappa))V^\top$ is implemented as $B$ plus a detached correction. For $\kappa>0$ it is positive definite with $\lambda_{\min}(|B|)\ge\eta\kappa$, hence
\begin{equation}
\kappa\,u^\top|B|^{-1}u\le\frac{\|u\|^2}{\eta}.
\tag{S1}
\end{equation}
Where $B\succ0$ and $\lambda_{\min}(B)\ge\eta\kappa$, the floor is inactive and $|B|=B$, so at $a=\ga$ the temporal term equals $2\kappa$ times the hold-regret of \cref{cor:regret} up to $O(\kappa\|\Delta s\|^3)$.
With the correction and the curvature scale detached,
\begin{equation}
\frac{\partial(\kappa u^\top|B|^{-1}u)}{\partial B}=-\kappa\,\gs\gs^\top,\qquad
\frac{\partial(\kappa u^\top|B|^{-1}u)}{\partial u}=2\kappa \gs .
\tag{S2}
\end{equation}
\Cref{tab:ablation} showed higher sm when the curvature was detached. With the correction and $u$ held fixed, a descent step on $B$ adds a positive multiple of $\kappa\gs\gs^\top$, which raises the curvature along the shift and leaves the orthogonal directions unchanged, and by the Sherman--Morrison formula this step lowers the energy $u^\top|B|^{-1}u$. With the scalar metric $\mathrm{tr}(B)/d$ of \cref{tab:ablation_full} the derivative is a multiple of the identity where the floor is inactive, so the same step raises the curvature equally in every direction. The anisotropic metric concentrates this local increase along the surrogate shift.
These derivatives describe independent variations of the local quadratic model. Training recomputes the correction and $u$ at every update and moves $B$ through shared parameters, so this local calculation does not establish a monotonic decrease of the recomputed energy or a monotonic increase of the curvature.

\subsection{Reward scale of the loss}
\label{supp:scale_prop}
\begin{proposition}[Value scale]\label{prop:scale}
For $Q\mapsto cQ$ with $c>0$, $u\mapsto cu$, $B\mapsto cB$, $\gs$ is invariant, and the unclamped batch scale $\kappa=\E\big[\tfrac1d\sum_i|\lambda_i(B)|\big]$ maps to $c\kappa$. Therefore, for the metric $|B|$ of \cref{eq:absB},
\begin{align}
\kappa u^\top|B|^{-1}u&\mapsto c^2\kappa u^\top|B|^{-1}u,\tag{S3}\label{eq:scale_npr}\\
\mathrm{relu}(v^\top\Haa v+\delta)&\mapsto c\,\mathrm{relu}(v^\top\Haa v+\delta/c).\tag{S4}\label{eq:scale_hinge}
\end{align}
\end{proposition}
\begin{proof}
Under $Q\mapsto cQ$ every derivative of $Q$ is multiplied by $c$ with unchanged eigenvectors, so $\kappa\mapsto c\kappa$ and $|B|\mapsto c|B|$. Hence $\gs=|B|^{-1}u$ is unchanged and $\kappa\,u^\top|B|^{-1}u$ is multiplied by $c\cdot c^2\cdot c^{-1}=c^2$. For the hinge, $\mathrm{relu}(ct+\delta)=c\,\mathrm{relu}(t+\delta/c)$.
\end{proof}
The implementation follows this identity when the numerical clamp is inactive before and after scaling.
The TD3 target with rewards $cr$ and critics $cQ_k$ is $c$ times the original target, so $\mathcal L_{\mathrm{TD}}\mapsto c^2\mathcal L_{\mathrm{TD}}$ and the loss of \npr{} has the same degree, whereas the fixed-margin hinge of PAVE does not. For SAC the same holds when the entropy temperature is scaled by $c$. The statement concerns the loss; it does not state that finite-step training follows a scaled trajectory or that the return is invariant.

\end{document}